%% file: main.tex
\documentclass[10pt,twocolumn,letterpaper]{article}

\usepackage{cvpr}              

\usepackage{epsfig}
\usepackage{graphicx}
\usepackage{amsmath}
\usepackage{amssymb}
\usepackage{booktabs}
\usepackage{epigraph}
\usepackage{url}            
\usepackage{amsfonts}       
\usepackage{nicefrac}       
\usepackage{xcolor,colortbl}         
\usepackage{xspace}
\usepackage{multirow}
\usepackage{mathtools}
\usepackage{overpic}
\usepackage{algpseudocode}
\usepackage{algorithm}

\usepackage{caption}

\usepackage{array}
\newcolumntype{H}{>{\setbox0=\hbox\bgroup}c<{\egroup}@{}}

\input{preamble}

\definecolor{cvprblue}{rgb}{0.21,0.49,0.74}
\usepackage[pagebackref,breaklinks,colorlinks,citecolor=cvprblue]{hyperref}

\def\paperID{439} 
\def\confName{CVPR}
\def\confYear{2024}

\title{\texttt{SeMoLi}: What Moves Together Belongs Together}

\author{Jenny Seidenschwarz$^{1,2}$\thanks{Correspondance to \tt\small j.seidenschwarz@tum.de.}
\quad
Aljoša Ošep$^{2}$
\quad
Franceso Ferroni$^{2}$
\quad
Simon Lucey$^{3}$
\quad
Laura Leal-Taix\'{e}$^{2}$\\
\\
$^1$\text{Technical University of Munich}
\hspace{1cm}
$^2$\text{NVIDIA}
\hspace{1cm}
$^3$\text{University of Adelaide}
}

\begin{document}

\twocolumn[{%
\renewcommand\twocolumn[1][]{#1}%
\maketitle
\begin{center}
\vspace{-0.5cm}
%
%
\includegraphics[width=0.99\linewidth]{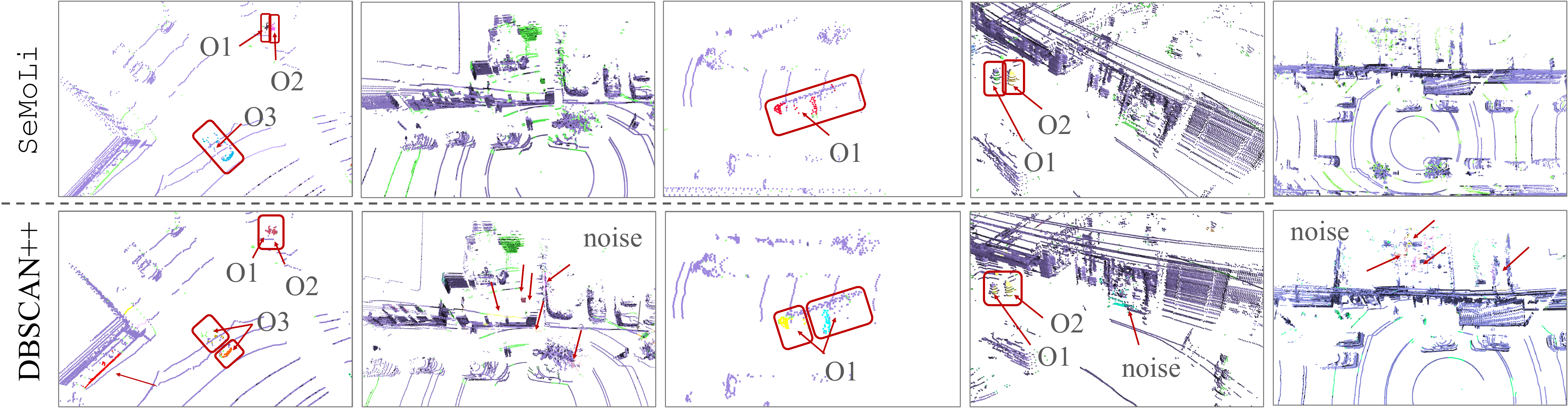}
\vspace{-0.1cm}
\captionof{figure}{\textbf{Towards learning to pseudo-label}: We propose \method, a data-driven approach for segmenting moving instances in point clouds (\textit{top}), that we utilize to learn to detect moving objects (O) in Lidar. We visually contrast \method to prior art, that tackles similar problem via density-based clustering (DBSCAN)~\cite{najibi2022motion}. 
%
We visualize the whole point cloud in purple, and dynamic points, used as input to our method and baseline to localize moving instances, in green. We color-code individual segmented instances. 
From left to right \method (i) segments objects even for sparse point clouds and suffers less from under-segmentation, (ii) is able to learn to filter noise from the filtered point cloud, (iii) leads to less over-segmentation, and (iv) generalizes better to different classes. \textit{Best seen in color, zoomed.}}
\end{center}%
}]
\input{sec/0_abstract}

\makeatletter{\renewcommand*{\@makefnmark}{}
\footnotetext{$^*$ Correspondence to j.seidenschwarz@tum.de.}\makeatother}

\input{sec/1_intro}

\input{sec/2_related}
\input{sec/3_method}

\input{sec/4_experiments}

\input{sec/5_conclusion}


{\small
\bibliographystyle{ieeenat_fullname}
\bibliography{refs}
}

\input{sec/X_suppl}


\end{document}

%% file: preamble.tex
\newcommand{\PAR}[1]{\vskip4pt \noindent {\bf #1~}}
\newcommand{\PARit}[1]{\vskip4pt \noindent {\it #1~}}

\newcommand{\method}{\texttt{SeMoLi}\@\xspace}
\newcommand{\methodfull}{\textit{Segment Moving in Lidar}\@\xspace}

\newcommand{\pipeline}{\texttt{DeMoLi}\@\xspace}

%% file: sec/0_abstract.tex
\begin{abstract}
\vspace{-6pt}
We tackle semi-supervised object detection based on motion cues. Recent results suggest that heuristic-based clustering methods in conjunction with object trackers can be used to pseudo-label instances of moving objects and use these as supervisory signals to train 3D object detectors in Lidar data without manual supervision. We re-think this approach and suggest that both, object detection, as well as motion-inspired pseudo-labeling, can be tackled in a data-driven manner. 
We leverage recent advances in scene flow estimation to obtain point trajectories from which we extract long-term, class-agnostic motion patterns. 
Revisiting correlation clustering in the context of message passing networks, we learn to group those motion patterns to cluster points to object instances. 
By estimating the full extent of the objects, we obtain per-scan 3D bounding boxes that we use to supervise a Lidar object detection network. Our method not only outperforms prior heuristic-based approaches ($57.5$ AP, $+14$ improvement over prior work), more importantly, we show we can pseudo-label and train object detectors across datasets. 
%
\end{abstract}

%% file: sec/1_intro.tex
\section{Introduction}


%
We tackle semi-supervised object detection in Lidar data in the context of embodied navigation. 




\PAR{Status quo.} The established approach to 3D Lidar object detection is to pre-define object classes of interest and train data-driven models with manually labeled data in a fully supervised fashion. Thanks to Lidar-centric labeled datasets~\cite{Geiger12CVPR,sun20CVPR,nuscenes2019} and developments in learning representations from unordered point sets \cite{qi2017pointnet,qi2017pointnet++}, the quality in Lidar-based detection continues to improve. 
However, this process that relies on manual labeling is notoriously expensive and, importantly, does not scale well to rare classes. 
Before the advent of end-to-end data-driven Lidar perception, pioneering works~\cite{Teichman11ICRA,Teichman12IJRR,Held14RSS,Held16RSS,Moosmann09IVS,Moosmann13ICRA,Zhang13} were inspired by the Gestalt principle \cite{wertheimer1922untersuchungen} of spatial proximity, \ie, the intuition that spatially close-by points belong together. To this end, these works utilize local, distance-based point clustering methods to segment objects via bottom-up grouping. They are class-agnostic and general, but inherently unable to benefit from the increasing amount of available data, and therefore, fall behind data-driven methods. 
Recently, heuristic approaches resurfaced within multi-stage pipelines for data auto-labeling in the context of semi-supervised 3D object detection~\cite{Osep19ICRA,najibi2022motion,zhang2023towards}. These auto-labeling pipelines (i) segment objects using local, bottom-up clustering methods, (ii) use Kalman filters to track segmented regions across time, and (iii) register segments across time to estimate amodal extents of objects to generate pseudo-labels for object detection training. While impressive, they still suffer from the same drawback: they are inherently unable to improve pseudo-labeling performance in a data-driven fashion. 

\PAR{Stirring the pot.} We re-think this approach and suggest that both object detection training based on pseudo-labels, as well as the pseudo-labeling process can be tackled in a data-driven manner. Prior efforts, too, leverage \textit{some} amount of labeled data to tune the hyper-parameters of the auto-labeling engine and consolidate design decisions. By contrast, we embrace a data-driven approach fully. We study how to train a pseudo-labeling network in a way that it can still be general and applied to an open-world setting as well as how its performance changes with different amounts of training data. 

\PAR{Learning to \textit{motion-}cluster.} Our approach \methodfull (\method) receives a set of Lidar point trajectories, obtained from a sequence of Lidar data. In contrast to prior heuristic approaches, we explore Gestalt principles to \textit{train} a Message Passing Network (MPN) \cite{gilmer2017MPN} with the objective of grouping points in a class-agnostic manner. Additionally to proximity, we leverage the intuition of \textit{common fate}, \ie, we learn to group points based on the observation that \textit{what moves together, belongs together} given \textit{some} labeled instances of moving objects. This way, we make motion a \textit{first-class citizen}. This, in turn, streamlines pseudo-labeling and alleviates the need for manual geometric clustering. 
We then use \method to identify motion patterns in unlabeled sequences and pseudo-label these at the per-point segmentation level. By extracting bounding boxes and inflating them to better match the amodal ground truth bounding boxes we train an off-the-shelf 3D detection network.

\PAR{Impact.} Our data-driven approach not only allows us to perform better as compared to a hand-crafted pipeline based on density-based clustering~\cite{najibi2022motion} -- more importantly, we demonstrate that \method's performance improves with increasing amount of data while still remaining a general approach applicable on different classes and datasets. Just as important, we democratize research of motion-inspired auto-labeling in Lidar streams by making our models, code, as well as previously inaccessible baselines~\cite{najibi2022motion} publicly available to foster future research.

To conclude, as our \textbf{main contribution} we (i) show that class-agnostic, motion-based segmentation for pseudo-labeling
can be tackled in a data-driven manner. To operationalize this idea, we (ii) propose a \textit{data-driven} point clustering method leveraging motion patterns that utilizes MPNs, constructs a graph over a Lidar point cloud, and learns which points \textit{belong together} based on spatial and motion cues. We (iii) demonstrate this network can be used to successfully pseudo-label unlabeled sequences even across unknown classes and datasets without the requirement of re-training. Finally, we (iv) make this field of research reproducible and establish a solid ground for making progress in this field of research as a community. We hope that this is a stepping stone towards a better world in which friendly robots do not accidentally crash into rare moving objects just because no one wants to label data manually.

%% file: sec/2_related.tex
\section{Related Work}
\label{sec:related}


\PAR{Lidar-based 3D Object Detection.}
Amodal 3D localization of objects is instrumental for situational awareness in dynamic environments and has been playing a pivotal role in autonomous navigation since its inception~\cite{thorpe1991toward, Petrovskaya09AR}. Fueled by the recent availability of high-quality Lidar-perception-centric datasets~\cite{Geiger12CVPR,Caesar20CVPR,sun20CVPR}, several methods directly learn a representations from unordered point sets~\cite{qi2017pointnet, qi2017pointnet++} to localize and classify objects~\cite{Qi17CVPR_frustum,Shi19CVPR,yang20203dssd,qi2019deep,liu2021iccv}, or utilize sparse 3D convolutional backbones~\cite{choy20194d,zhou2018voxelnet,yan2018second}. Several methods \cite{yang18cvprPixor,yin2021center} flatten 3D voxel representation along the height dimension, while~\cite{Lang19CVPR} replaces regular grid-based discretization with vertical columns. 
%
%
All aforementioned methods rely on manually labeled amodal 3D bounding boxes that encapsulate the full extent of the object. Manual annotations of such regions are costly, and, for practical purposes, constraint \textit{target} classes that must be detected to a fixed, pre-defined set of objects.

\PAR{Lidar-based bottom-up grouping.} Prior to the advent of end-to-end object detection models, Lidar perception models have been relying on bottom-up point grouping~\cite{douillard2011segmentation}. Several methods utilize the connected components algorithm on graphs, implicitly defined over the rasterized birds-eye-view representation of point clouds~\cite{Teichman11ICRA,klasing2008clustering} or the full point cloud~\cite{Wang12ICRA} to localize disconnected regions, \ie, objects in Lidar scans. 
%
%
Instead of utilizing graph-based representations, several methods~\cite{Behley13IROS,Mcinnes2017OSS,hu2020learning,wong2020identifying,najibi2022motion,nunes2022unsupervised} utilize robust density-based clustering methods (DBSCAN)~\cite{Ester96KDD} to group points. 
The advantage of these methods is that the segmentation is class-agnostic, largely relying on the Gestalt principle \cite{wertheimer1922untersuchungen} of spatial proximity, which is a strong cue in the Lidar domain. 
However, these methods do not benefit from the increasing amount of (labeled) data to improve segmentation performance. Our work is inspired by prior efforts that localize objects as connected components in graphs~\cite{Teichman11ICRA,Wang12ICRA}. We revisit these ideas in the context of data-driven grouping while still remaining class-agnostic behavior.

\begin{figure*}[htp!]
    \centering
    \includegraphics[width=0.9\linewidth]{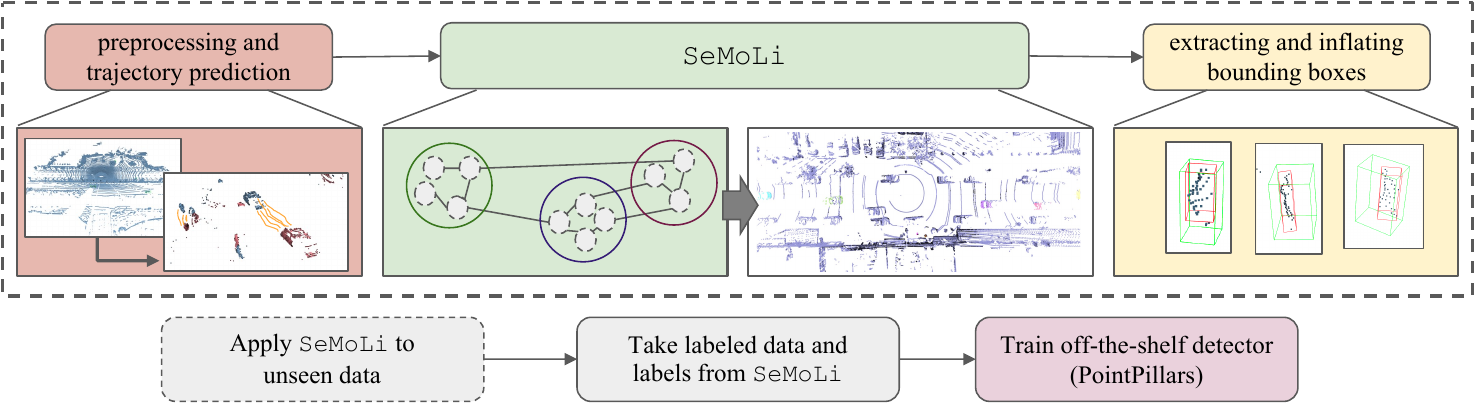}
    \caption{\textbf{Segment Moving in Lidar for Pseudo-Labeling:} We first preprocess the point cloud to remove static points and predict per-point trajectories on the filtered point cloud (\textit{preprocessing and trajectory prediction}). Then, we extract velocity-based features from the trajectories and \textit{learn} to cluster, \ie, segment points based on motion-patters using a Message-Passing Netowrk \cite{gilmer2017MPN} in a fully data-driven manner (\method). From point segments, we extract bounding boxes and inflate them (\textit{extracting and inflating bounding boxes}). Finally, we apply our approach on \textit{unlabeled} Lidar streams to obtain pseudo-labels, that we use to train object detectors.}
    \label{fig:overview}
\end{figure*}

\PAR{Motion-supervised object detection in Lidar.} 
Tracking-based semi-supervised learning in Lidar is investigated in~\cite{Teichman12IJRR} by performing bottom-up segmentation of Lidar point clouds, followed by Kalman-filter-based object tracking of Lidar segments to expand the core labeled training set via tracking. A recent work by~\cite{najibi2022motion} follows a similar pipeline, however, first filters out stationary points to obtain supervision for moving objects, as needed to train object detectors. Another related approach by~\cite{zhang2023towards} follows a similar pipeline,
and additionally proposes a novel data augmentation that allows object detectors to generalize to far ranges solely based on instances observed in near sensing areas. Unfortunately, the aforementioned are proprietary and do not share models, experimental data, or evaluation/implementational details. As these consist of hand-crafted pipelines, results reported by these methods are difficult to reproduce, hindering the community's progress in this field of research.

%
%
%
 
%

We tackle a similar problem, however, by contrast, we propose a data-driven, graph-based model that learns a representation of the point cloud, suitable for localizing moving objects in Lidar. Our method performs favorably compared to state-of-the-art even when trained on a tiny amount of labeled data, and further improves with an increased amount of training data. Moreover, we show we can apply our approach to other datasets without any re-training of our pseudo-labeling model. To foster future research, we make our models, code, and experimental data publicly available. 

%% file: sec/3_method.tex
\vspace{+0.1cm}
\section{Learning Moving Objecs in Lidar}
\label{sec:method}

We present \methodfull (\method), our data-driven approach for motion-inspired pseudo-labeling for 3D object detection. Our \textit{teacher} network learns to cluster points in Lidar point clouds based on their motion patterns (\cref{sec:learning_to_cluster}). This step can also be seen as point cloud segmentation and corresponds to localizing moving object instances. Once trained, we utilize \method to segment \textit{unlabeled} Lidar data, estimate amodal 
object-oriented bounding boxes from the segments and use these as a supervisory signal to train a \textit{student} 3D object detector (\cref{sec:object_detection}). 


\subsection{Learning to Segment Moving Objects}
\label{sec:learning_to_cluster}

Given input Lidar point clouds $\mathcal{P}$, our task is to localize individual moving objects as 3D bounding boxes. Each object will be represented by an a-priori unknown number of points, which we group together based on cues derived from Gestalt principles~\cite{wertheimer1922untersuchungen}. We propose to use \textit{proximity}, \ie, nearby points likely belong to the same object, and \textit{common fate}, \ie, what moves together belongs together.
Prior works cluster points using, \eg, density~\cite{Ester96KDD,najibi2022motion,wang2019towards}, or graph-based clustering methods~\cite{Wang12ICRA} that base grouping decisions on the local context and do not leverage the benefits of learning-based algorithms. 
%
This is problematic for Lidar. On the one hand, signal sparsity and the vast range of object sizes that appear in the dataset (\eg, pedestrian \vs truck) often lead to over- and under-segmentation~\cite{Held16RSS}. On the other hand, the performance of such approaches suffers heavily in the presence of background noise.

\PAR{Correlation clustering revisited.} 
To segment point clouds into a set of instances, we lean on correlation clustering~\cite{bansal2004correlation}. We represent a point cloud as a weighted graph $G = (V, E)$ with nodes $V$ and edges $E$. A node $n_i$ represents a point and encodes its spatial position and motion as node features $h_i^{(0)}$ while an edge $e_{i,j}$ represents the geometrical connection between two points $i$ and $j$ with edge features $h_{ij}^{(0)}$ representing their relationships. 
The clustering algorithm then cuts edges to obtain a set of connected components which represent point cloud instance segmentations. 
We revisit correlation clustering in the context of Message Passing Networks (MPNs)~\cite{gilmer2017MPN}. Hence, we define node and edge features in a data-driven manner to \textit{learn} edge scores given \textit{some} labeled data. 
%

\PAR{Talk to your neighbors.} 
MPNs propagate information across the graph and ensure that the \textit{learned} graph partitioning does not \textit{only} rely on the local relationships between points. Initial node $h_i^{(0)}$ and edge features $ h_{i,j}^{(0)}$ are updated in an iterative manner over $L$ layers. This ensures that the final edge features $h_{i,j}^{(L)}$ contains global information, providing the necessary context needed to decompose the graph into object instances. We utilize a binary edge classifier on top of $h_{i,j}^{(L)}$ to obtain our edge scores. In the following, we discuss how we construct and parameterize our input graph, and how we learn the representation using a MPN.

\subsubsection{Graph Construction}
\label{subsec:graph_construction}


\PAR{Point cloud sequence preprocessing.} 
Given a Lidar point cloud sequence $\mathcal{P} = \{ P^t \in \mathbb{R}^{N^t\times 3} \}, t \in 1,\ldots,T$. 
We first remove static points from the raw point clouds in $P^t$.
This yields a sequence of stationary point clouds $\tilde{\mathcal{P}} = \{ \tilde{P}^t \in \mathbb{R}^{M^t\times 3} \}, t \in 1,\ldots,T$, 
where $M^t \leq N^t$. Our pre-filtering step closely follows prior work~\cite{Dewan15ICRA,najibi2022motion} (see appendix for details). For simplicity, we omit index $t$ in the remainder of the paper. 

\PARit{Motion cues.} For each filtered point cloud $\tilde{P}$ we predict ego-motion compensated point trajectories $\mathcal{T} \in \mathbb{R}^{M\times (3\times24)}$ using a self-supervised trajectory prediction network~\cite{wang2023NFT}. Each point trajectory $\tau_i \in \mathcal{T}$ is defined as a sequence of point positions $\tau_i = \{p_i^k\}_{k=0}^{24}$. We utilize trajectory information to encode motion cues into our graph features, as described below.


\PAR{Graph nodes $V$.} 
For each filtered point cloud $\tilde{P}$, we view points as nodes in a graph with corresponding node features. 

\PARit{Node parametrization.}
The node feature $h_i^{(0)}$ includes its spatial coordinates $(x, y, z)$ and the statistical measures (mean, min, max) of the velocities along its trajectory:
\begin{equation}
    h_i^{(0)} = \left( x_i, y_i, z_i, \text{mean}(v_{\tau_i}), \text{min}(v_{\tau_i}), \text{max}(v_{\tau_i}) \right).
\end{equation}
These features capture both the spatial position and the dynamic behavior of each point. The velocity at each time step is calculated as the difference between consecutive points in the trajectory: $v_{\tau_i} = \left\{ p_i^{k+1} - p_i^k \right\}_{k=0}^{k=24}$.





\PAR{Graph edges $E$.} We connect nodes $n_i \in V$ with edges $e_{i,j}$ that hypothesize point-to-instance memberships. 
%
%
Utilizing a fully connected graph is computationally infeasible and introduces a large amount of negative edges that impedes the learning process.
Hence, we leverage proximity as an inductive bias based on the intuition that ``far away'' points are unlikely to belong to the same object instance. We constrain node connectivity to a set of \textit{k}-nearest neighboring nodes in terms of Euclidean distance. We discuss alternative graph construction strategy in the experimental section (\cref{sec:ablations}). Note that the final edge set is still biased towards negative edges. 
%

\PARit{Edge parametrization.} We parametrize edges $e_{ij}$ connecting nodes $i$ and $j$ via initial edge features as the difference in their spatial coordinates: 
\begin{equation}
    h_{ij}^{(0)} = \left( x_i - x_j, y_i - y_j, z_i - z_j \right)
\end{equation}
This captures the relative spatial relationship between the points in the point cloud. 
To ensure each edge obtains a global view of the point cloud, as needed for reliable clustering, we perform several Message Passing iterations, as described below. 

\subsubsection{Representating Learning via Message Passing}

The message passing algorithm involves iteratively updating node and edge embeddings. For a fixed number of iterations $L$, we perform updates as follows:

%

\PAR{Edge update.} 
At each step $l \in \{1, \dots, L\}$, we update the embedding of an edge $e_{ij}$ connecting nodes $i$ and $j$ based on its previous embedding $h_{ij}^{(l-1)}$ and the embeddings of the adjacent nodes $h_i^{(l-1)}, h_j^{(l-1)}$: 
\begin{equation}
    h_{ij}^{(l)} = f\left(h_{ij}^{(l-1)}, h_i^{(l-1)}, h_j^{(l-1)}\right),
\end{equation}
where $f(\cdot, \cdot, \cdot)$ is an shared-weight update function, that consists of a linear, a normalization, and a dropout layer.

\PAR{Node update.} 
Similarly, we update node embeddings based on their previous embeddings as well as their neighbors previous embeddings:
\begin{equation}
m_{i,j}^{(l)} = g\left(h_{ij}^{(l)}, {h_j^{l-1}, h_i^{(l-1)}}\right),
    h_{i}^{(l)} = \phi(\{m_{i,j}^{(l)}\}_{j \in N_i}),
\end{equation}
where $g(\cdot, \cdot)$ is the node update function with shared weights over all layers that constitutes of a linear layer, a normalization layer, and a dropout layer, $\phi$ is a mean aggregation function and $N_i$ denotes the set of nodes adjacent to node $i$. The iterative process of updating node and edge embeddings allows for the integration of local and global information in the graph, enabling the algorithm to capture complex patterns and relationships within the data.

\PAR{Edge classification.}
To determine the final edge score, we feed the final edge features $h_{ij}^{(L)}$ through a final linear layer $f_f$ followed by a sigmoid layer $\sigma$:
\begin{equation}
    \Tilde{h}_{ij}^{(L)} = \sigma\left(f_f(h_{ij}^{(L)})\right).
\end{equation}

\noindent For training details we refer the reader to appendix.

\PAR{Postprocessing.}
During inference, we first cut negative edges with score $\Tilde{h}_{ij}^{(L)} < 0.5$. To ensure robustness towards a small set of possibly miss-classified outlier edges, we apply correlation clustering~\cite{bansal2004correlation} using our learned edge scores (details in the appendix) on top of this graph. We discard all singleton nodes without edges. The resulting point clusters $c_c \in C_c$ represent our segmented object instances. 

\subsection{Learning to Detect Moving Objects}
\label{sec:object_detection}

Training the student network requires to, first, transform a segmented point cluster $c_c$ to a bounding box $b_c$. Then we enhance them by inflation to obtain our final pseudo-labels. 


\PAR{From point clusters to bounding boxes.}
Given a set of points that constitutes a point cluster $c^c \in C_c$, we determine the translation vector $t_c$ by taking the midpoint of all points. 
Since each point $i$ has a trajectory $\tau_i$ assigned to it, we can compute the mean trajectory $\hat{\tau_c} = \{\hat{p}_c^k\}_{k=0}^{24}$ and leverage $\hat{p}_c^1$ and $\hat{p}_c^2$ to determine the heading of the object in $xy$-direction $\alpha_c$.
With the heading $\alpha_c$ and the translation vector $t_c$ we can transform the points to determine axis-aligned length, width, and height of the bounding box $lwh_c$ which yields the 3D bounding box $b_c = [t_c, lwh_c, \alpha_c]$. 

\PAR{Bounding Box Inflation.}
The bounding boxes $b_c$ represent the most compact enclosing axis-aligned cuboid. Ground truth bounding boxes are represented by typically more loose, amodal bounding boxes. To adapt $b_c$ to the corresponding ground truth data, we inflate them to have a minimum length, width, and height of $x_{min}$, $y_{min}$, and $z_{min}$, respectively, to obtain our final pseudo-labels.

%
%

\PAR{Detector training.}
We train PointPillars~\cite{lang2019PP} (PP) two-stage object detector in a class-agnostic setting using our generated pseudo-labels. We utilize the detector implementation as well as hyperparameters settings from~\cite{mmdet3d2020} (see appendix), and perform the following changes. To account for objects of various sizes, we adapt anchor box generation with various size parameters (see appendix for details). We adapt the detection region to a planar $100x40m$ field, centered at the egovehicle (\cf, \cite{najibi2022motion}). We use binary cross-entropy loss to train our object classifier (\texttt{object}, \texttt{background}). 


%% file: sec/4_experiments.tex
\section{Experimental Validation}
\label{sec:experiments}

In this section, we thoroughly ablate our motion-inspired, data-driven approach \methodfull (\method) and its individual components. 
In Sec.~\ref{sec:experimental_setup}, we outline our evaluation test-bed, which we use in Sec.~\ref{sec:ablations} to discuss several design decisions behind our pseudo-labeling network. In Sec.~\ref{sec:semi_supervised_3d_object_detection}, we evaluate the performance of our object detector, trained using our pseudo-labels.

\subsection{Experimental Setup}
\label{sec:experimental_setup}

\PAR{Datasets.} We evaluate our method using Waymo Open dataset~\cite{sun20CVPR} and only utilize Lidar data. The dataset provides labels as amodal 3D bounding boxes for \texttt{pedestrian, vehicle} and \texttt{cyclist} classes. 
We assess the generalization of our method on Argoverse2 dataset~\cite{wilson2021argoverse2} which provides finer-grained semantic labels. Both datasets were recorded with different types of (proprietary) Lidar sensors (for details we refer to the appendix). 

\PAR{Evaluation setup.} Following \cite{najibi2022motion}, we evaluate \method as well as the final object detector in a $100x40m$ rectangular region (region of highest importance for autonomous vehicles), centered in the ego-vehicle Lidar sensor center. 
%
As our pseudo-labels and trained detectors do not provide fine-grained semantic information, we follow the literature~\cite{najibi2022motion,zhang2023towards} and evaluate both, \method and \pipeline in a class-agnostic setting. 
We assess our object detectors on \texttt{moving-only} (\cf, \cite{najibi2022motion}) as well as \texttt{all} labeled objects to see how well training instances mined from moving regions generalize to non-moving objects. 
%
%
We consider an object as moving if its velocity is larger than $1 m/s$.
We treat \texttt{non-moving} instances as ignore regions, \ie, we ignore detections having \textit{any} intersection with static objects, consistent with the methodology used in prior works~\cite{najibi2022motion}. 
%
%
\PAR{Semantic Oracle.} For per-class recall analysis we assign labels to class-agnostic detections if they have any 3D IoU overlap with labeled boxes. 


\begin{figure}[t]
    \centering
\includegraphics[width=0.95\linewidth]{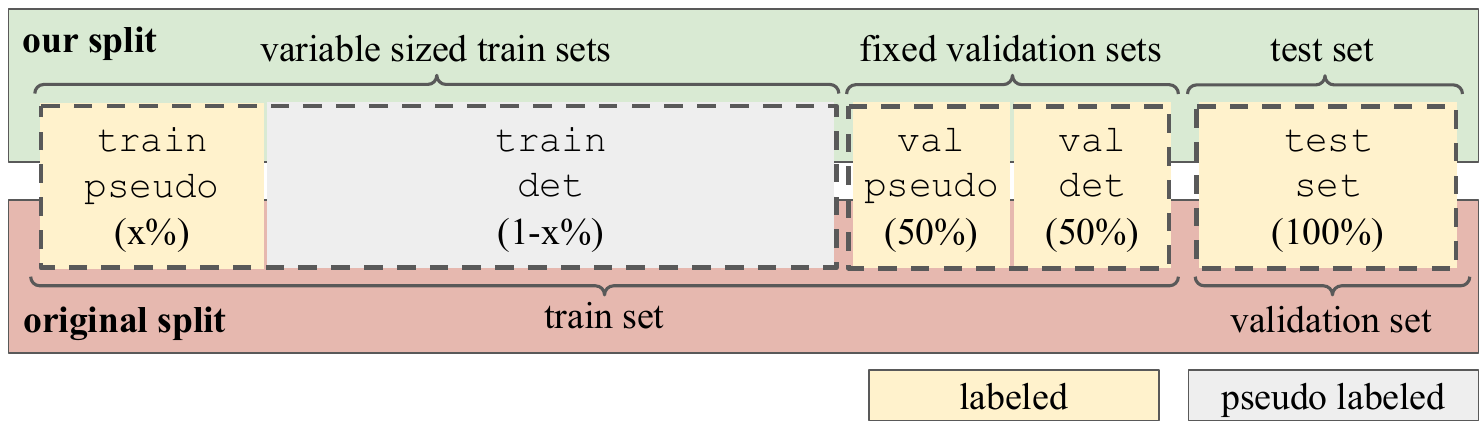}
\vspace{-0.3cm}
    \caption{\textbf{Train and validation splits:} We conduct our experiments using Waymo training set, for which manual labels are available. We pre-fix two separate validation sets, one for validating pseudo-labels (\texttt{val\_pseudo}), and one for end-model detector performance (\texttt{val\_det}). We report performance on varying ratios $x$ for training \method (\texttt{train\_pseudo}) and generating pseudo-labels for training our detector (\texttt{train\_det}).}
    \label{fig:train_val_splits}
\end{figure}

\PAR{Metrics.} To evaluate 
\textit{pseudo-label generation}, we report the F1 score, defined as the harmonic mean of precision and recall. This metric is suitable for evaluating a set of predictions that are not ranked, as is commonly employed in the segmentation community~\cite{Kirillov19CVPR}. As the output of this step is point cloud instance segmentation, we report results using both, the mask intersection-over-union (SegIoU) criterion, as well as 3D bounding box IoU (3DIoU) criterion for quantifying true positives and false negatives.
%
%
This is important as it assesses how well we can recover labeled amodal bounding boxes that we need to use to train object detectors. For \textit{Object detection}, the evaluate the performance using the standard Average Precision metric that assumes as input a ranked set of object detections. We focus on 3D bounding box IoU as an evaluation criterion. 
%
%
For pseudo-label generation as well as object detection, we report performance using localization thresholds of $T\leq\{0.7, 0.4\}$ with $0.7$ being the standard evaluation threshold on Waymo Open dataset and $0.4$ the threshold used in \cite{najibi2022motion}.

\PAR{Data splits.} 
Each approach for pseudo-labeling requires a certain percentage of labeled data to tune hyper-parameters. We embrace a fully data-driven approach to pseudo-labeling: given \textit{some} labeled data, we train \method to localize moving objects in Lidar sequences. We then pseudo-label the unlabeled set of training data, which we use to train our \textit{student} object detection network. 
This requires a careful evaluation protocol, outlined in \cref{fig:train_val_splits}. 
We split the official Waymo training set into fixed-size training (\texttt{train}) and validation (\texttt{val}) splits. 
%

\noindent \texttt{\textbf{Val}}. As we need to train and validate \textit{teacher} and \textit{student} models independently, we split the \texttt{val} set into two equal-sized splits. We use \texttt{val\_pseudo} for the validation of the \textit{teacher} network, and \texttt{val\_det} to independently validate the \textit{student} (detection) network (Fig.~\ref{fig:train_val_splits}). 

\noindent \texttt{\textbf{Train}}. We split the \texttt{train} set into (labeled) training set \texttt{train\_pseudo}, used to train the teacher network, then we utilize this network to pseudo-label the \textit{student} training set \texttt{train\_pseudo}. We use pseudo-labels to train the detection network. 
Note that we split the dataset along sequences, \ie
we do not allow frames of the same sequence to be in different splits.

\PAR{Varying train set size.} To study the amount of data needed to train our model, we sample varying-sized subsets for training the \textit{teacher} and \textit{student} networks, \ie, \texttt{train\_pseudo} and \texttt{train\_det}. 
Consistent with prior work~\cite{najibi2022motion}, we use the official Waymo Open \textit{validation} for the end-model evaluation. This set provides labels and allows us to focus our evaluation on moving objects.

\vspace{-0.1cm}
\subsection{Ablations Studies}
\label{sec:ablations}
\vspace{-0.1cm}


This section ablates the impact of different graph construction as well as parametrization strategies and bounding box inflation on \texttt{val\_pseudo} dataset. Features based on velocity and position are defined as in \cref{subsec:graph_construction}.

\PAR{Graph construction strategies:} As constructing a fully connected graph is not feasible, we investigate different strategies to construct a k-nearest-neighbor (kNN) graph (\cref{subsec:graph_construction}) based on distance and velocity as similarity measures.
In \cref{tab:gnn_ablation}) we report the oracle performance for both, \ie, the performance we achieve with each strategy assuming all edges are classified correctly. We base our design choice of utilizing position on its superior oracle performance ($90.9$$@0.4$ F1) (see appendix for deeper analysis).
%

%
%

\input{tables/ABLATIONS_GNN}
\input{tables/ABLATION_INFLATION}
\input{tables/PLQ}

\PAR{Node features:} As we show in \cref{tab:gnn_ablation}, combining velocity and position in our node features significantly aids the learning process.
%
Solely utilizing velocity lowers recall and slightly increases precision compared to position-based features, indicating an increased amount of rejected positive edge hypotheses due to possibly noisy velocity predictions. 
On the other hand, position as node features leads to a severe drop in precision and a slight drop in recall indicating that \method fails to correctly classify negative edges. Concatenating both, velocity enables \method to reject negative edges between points in close proximity if they do not move together and clusters points \textit{that move together}.

\PAR{Edge features:} As in graph construction, we consider $L2$ distance between position- and velocity-based features, as well as the concatenation of both. 
As we show in \cref{tab:gnn_ablation}, adding velocity to or completely omitting position from edge features harms the performance. Therefore, we use position-based encoding for edges. 


\PAR{Bounding Box Inflation}
We inflate our bounding boxes to a minimum length, width, and height and show the impact on the performance on Waymo Open Dataset in \cref{tab:ablation_amodalization}. While the performance based on SegIoU does not change significantly, evaluating the performance based on 3DIoU improves drastically. \method segments points correctly, but generates significantly more tight bounding boxes compared to ground truth (see appendix for details). 

\subsection{Evaluation of Pseudo-Label Quality}
\vspace{-0.2cm}
\label{sec:pseudo_label_quality}
\PAR{Baselines.} We compare \method to three baselines: vanilla DBSCAN, DBSCAN++ (\cf, \cite{najibi2022motion}), and its variant DBSCAN++$_{l}$ that utilizes our (long-term) velocity-based motion feature for a fair comparison. 
%
$^{\dagger}$ indicates heuristic filtering based on bounding box dimensions (detailed in the appendix). We also report variant of our method that utilizes local motion cues (scene flow), \method$_{\texttt{sf}}$.
\input{tables/PLQ_CLASS_WISE_WAYMO}

\PAR{Discussion.} We discuss results in terms of 3DIoU, as the output of this step is used to train object detectors that assume amodal bounding boxes. In \cref{tab:pseudo_quality} (\textit{top}) we report results we obtain with ground-truth motion information (oracle). In this setting, both DBSCAN ($69.0@0.4$ F1) and \method ($73.5@0.4$ F1) perform well.
%
%
%
However, when utilizing, estimated motion information, all DBSCAN variants struggle with precision -- even the variant with outlier removal is unable to surpass $16.2@0.4$ F1. 
By contrast, \method can learn to filter noise and undergoes a significantly less severe performance drop as compared to its motion oracle. All \method variants surpass $50@0.4$ F1, outperforming DBSCAN in terms of precision \textit{and} recall.  
\input{tables/PLQ_CLASS_WISE_ARGO} 
\PAR{Long-term motion cues.} As can be seen, by contrast to DBSCAN, our \method$^{\texttt{10}}$ method can learn to utilize long-term motion cues, and performs favorably compared to the variant that relies on local motion estimates, \method$^{\texttt{10}}_{\texttt{sf}}$. 

\input{tables/PP_Waymo}

\input{tables/PP_AV2}

\PAR{More data.} Finally, we show \method benefits from the increased amount of training data. \method's performance monotonically increases with expanding training set, eventually reaching $57.6@0.4$ F1. 




\PAR{Per-class analysis.}
%
In \cref{tab:class_wise} we discuss the per-class analysis (based on our oracle classifier, see \cref{sec:experimental_setup}). While we observe similar recall among \method and DBSCAN++$^{\dagger}$, we note that the latter 
%
%
produces significantly more false positive pseudo-labels, quantified via $\%$uFP, the percentage of pseudo-labels not matched to any ground truth object. We conclude that our approach produces a pseudo-label set that has a significantly better signal-to-noise ratio.


\PAR{Cross-dataset evaluation.}
We further examine the ability of cross-dataset generalization by evaluating our model \method$_{\texttt{90}}$ on the Argoverse2 dataset.
Importantly, we \textit{never} our \method on Argoverse2, but solely on the \texttt{train\_pseudo} split of Waymo Open dataset with $x=90\%$. Argoverse2 contains finer-grained semantic labels compared to Waymo, which allows us to assess generalization to a wide range of classes. We note that Waymo and Argoverse2 are recorded with different, proprietary Lidar datasets, requiring our approach to generalize across sensors. In \cref{tab:av2_performance} we show, \method again consistently performs favorably compared to the DBSCAN baseline. 




\subsection{Object Detection}
\label{sec:semi_supervised_3d_object_detection}
Finally, we utilize pseudo-labeled data to train an off-the-shelf object PointPillars (PP) object detector~\cite{lang2019pointpillars}. 

\PAR{Experimental setting.} As a baseline, we train PP in the standard setting with fine-grained semantic information (\textit{class-specific}) as well as \textit{class-agnostic}. The latter more closely resembles our setup, in which we do not have access to semantic information. Further, we train PP using all objects (\textit{stat. + mov}), as well as moving-only (\textit{mov-only}). 
We evaluate detectors on \textit{all} objects (\cref{tab:PointPillars}, \textit{top}), as well as only on \textit{moving} objects (\cref{tab:PointPillars}, \textit{bottom}, as reported in~\cite{najibi2022motion}).
%
Additionally, we mix pseudo-labeled with 10\% of labeled data (for finer-grained analysis see appendix) by re-using the $\texttt{train\_pseudo}$ splits (\cref{sec:experimental_setup}). 

%
\PAR{All objects.} 
In \cref{tab:PointPillars} (\textit{top}) we discuss results on \textit{all}, \ie, static and moving objects. We focus this discussion on $0.4$ threshold. As can be seen, with \textit{vanilla} PP detector, we obtain $80.3$ mAP. In the class agnostic setting, we obtain $64.7$ AP, which drops to $35$ AP when only training with GT boxes labeled as moving. When using 10\% of GT labels, we obtain $52.0$ and $54.8$ AP with DBSCAN and \method, respectively. Remarkably, when utilizing \textit{any} pseudo-labels in conjunction with 10\% labeled data, we obtain higher AP compared to the variant, trained with moving-only GT labels. This is likely because due to the noisy estimated flow, we retain \textit{some} static objects. The moving-only GT version learns to only predict objects in regions where moving objects are likely to appear. Additionally, pseudo-labels may induce generalization to the training process.
%
When not utilizing any labeled data, we obtain $14.9$ and $19.5$ AP, respectively. With \method, we, therefore, recover $56.6\%$ performance of the variant, trained on GT moving-only labels. 

%

\PAR{Moving objects.} 
When analyzing the performance on moving objects only in \cref{tab:PointPillars} (\textit{bottom}), models, trained on pseudo-labeled bounding boxes, are significantly closer to fully supervised models ($88.7$ trained on all data, and $89.0$ when trained with moving only). Utilizing no labeled data, DBSCAN reaches $43.2$ ($48 \%$ of GT model), while \method reaches $57.5$ ($64 \%$ of GT model). When using 10\% of labeled data, \method reaches $66.2$ AP ($64 \%$ of GT model). 
%
%
We note \cite{najibi2022motion} reports $40.4$ mAP on the validation set (last row), however, this is not an apple-to-apple comparison, as \cite{najibi2022motion} does not provide implementation or detailed description of how the analysis was conducted. Moreover, it is unclear \textit{how} \cite{najibi2022motion} reports mAP in a class-agnostic setting.



\PAR{Cross-dataset performance.} Finally, in \cref{tab:PointPillars_AV2} we report results we obtain by training PP detector on Argoverse2 dataset. We compare the supervised detector (\textit{labeled}) to \method, trained via pseudo-labels, generated on Argoverse2. Importantly, we generate pseudo-labels using \method trained on Waymo dataset, thus truly assessing cross-dataset generalization. 
When reporting results on \textit{all} objects, we obtain $35.5$ AP with the supervised model, and $22.9$ with ours. For moving objects, we obtain $82.9$ with the supervised model, and $57.6$ with our approach, confirming our approach is indeed general and transferable across datasets. We detail the training in appendix.


%% file: tables/ABLATIONS_GNN.tex
\begin{table}[t]
    \centering
    \footnotesize
    \setlength{\tabcolsep}{2.5pt}
    \begin{tabular}{ll|ccc|cccHHHHHH} 
    \toprule
        & Method & Pr 0.7 & Re 0.7 & F1 0.7 & Pr 0.4 & Re 0.4 & F1 0.4 & Pr 0.7 & Re 0.7 & F1 0.7 & Pr 0.4 & Re 0.4 & F1 0.4\\
        \midrule
        \midrule
        \parbox[t]{2mm}{\multirow{2}{*}{\rotatebox[origin=c]{90}{\textit{\textbf{Graph}}}}} & \textbf{Oracle} Velocity kNN & 35.7 & 67.1 & 46.6 & 39.4 & 74.1 & 51.4 & 4.2 & 7.9 & 5.5 & 16.3 & 30.7 & 21.2 \\
        & \textbf{Oracle} Position kNN & 85.0 & 87.9 & 86.4 & 89.4 & 92.5 & 90.9 & 9.7 & 10.0 & 9.8 & 37.4 & 38.7 & 38.0 \\
        \midrule
        \parbox[t]{2mm}{\multirow{3}{*}{\rotatebox[origin=c]{90}{\textit{\textbf{Node f.}}}}} & Velocity  & 61.2 & 52.7 & 57.0 & 73.0 & 62.2 & 67.1 & 6.5 & 5.6 & 6.0 & 31.6 & 26.9 & 29.0 \\
        & Position  & 57.6 & 61.7 & 59.6 & 65.2 & 69.9 & 67.5 & 6.6 & 7.1 & 6.8 & 28.8 & 30.9 & 29.8  \\
        & Velocity + Position & 69.4 & 58.0 & 63.2 & 77.9 & 65.1 & 70.9 & 8.1 & 6.7 & 7.3 & 33.2 & 27.8 & 30.3 \\
        \midrule
        \parbox[t]{2mm}{\multirow{3}{*}{\rotatebox[origin=c]{90}{\textit{\textbf{Edge f.}}}}} & Velocity & 58.9 & 48.6 & 53.2 & 69.6 & 57.4 & 62.9 & 6.0 & 5.0 & 5.4 & 27.2 & 22.4 & 24.5 \\
        & Position & 69.4 & 58.0 & 63.2 & 77.9 & 65.1 & 70.9 & 8.1 & 6.7 & 7.3 & 33.2 & 27.8 & 30.3  \\
        & Velocity + Position & 68.2 & 57.1 & 62.2 & 77.8 & 65.1 & 70.9 & 8.2 & 6.8 & 7.4 & 32.4 & 27.1 & 29.5  \\
    
    \bottomrule
    \end{tabular}
    \caption{
    \textbf{\method$^{\texttt{10}}$ ablation (SegIoU):} We discuss different strategies on \method \textbf{graph construction}, as well as \textbf{edge} and \textbf{node feature} parametrization.
    } 
    \label{tab:gnn_ablation}
\end{table}

%% file: tables/ABLATION_INFLATION.tex
\begin{table}[t]
    \centering
    \resizebox{1.0\linewidth}{!}{
    \footnotesize
    \setlength{\tabcolsep}{5pt}
    \begin{tabular}{l|ccc||ccc}
    \toprule
        & \multicolumn{3}{c}{\textit{3DIoU}} & \multicolumn{3}{c}{\textit{SegIoU}} \\
          & Pr 0.4 & Re 0.4 & F1 0.4 & Pr 0.4 & Re 0.4 & F1 0.4 \\
        \midrule
        Initial & 33.2 & 27.8 & 30.3 & 77.9 & 65.1 & 70.9 \\
        Inflated & 59.1 & 48.3 & 53.1 & 80.9 & 66.2 & 72.8\\
    \bottomrule
    \end{tabular}
   }
    \caption{\textbf{Bounding box inflation:} We inflate tight bounding boxes that enclose point clusters to a minimum width, length, and height. The segmentation performance changes only insignificantly while the detection performance improves drastically. \method$^{\texttt{10}}$ clusters points together correctly, but generates bounding boxes that are significantly tighter around the objects.} 
    \label{tab:ablation_amodalization}
\end{table}

%% file: tables/PLQ.tex
\begin{table}[t]
    \centering
    \footnotesize
    \setlength{\tabcolsep}{4.2pt}
    \begin{tabular}{l|ccc|ccc}
    \toprule
        Method & Pr 0.7 & Re 0.7 & F1 0.7 & Pr 0.4 & Re 0.4 & F1 0.4 \\
        \midrule
        \midrule
        \multicolumn{7}{l}{\textit{Oracle pseudo-label quality with ground truth flow / trajectories}}\\
        \midrule
        \textcolor{gray}{DBSCAN++} & \textcolor{gray}{20.8} & \textcolor{gray}{19.1} & \textcolor{gray}{19.9} & \textcolor{gray}{70.7} & \textcolor{gray}{64.9} & \textcolor{gray}{67.7} \\
        \textcolor{gray}{DBSCAN++$^\dagger$} & \textcolor{gray}{20.6} & \textcolor{gray}{19.1} & \textcolor{gray}{19.8} & \textcolor{gray}{72.0} & \textcolor{gray}{66.2} & \textcolor{gray}{69.0} \\
        \hline
        \textcolor{gray}{\method$^{\texttt{10}}$ } &  \textcolor{gray}{24.7} & \textcolor{gray}{23.1} & \textcolor{gray}{23.9} & \textcolor{gray}{76.0} & \textcolor{gray}{71.2} & \textcolor{gray}{73.5} \\ 
        \midrule
        \midrule
        \multicolumn{7}{l}{\textit{Pseudo-label quality with our computed flow / trajectories}}\\
        \midrule
        DBSCAN &  0.9 & 5.7 & 1.5 & 6.0 & 39.3 & 10.4 \\
        DBSCAN$^{\dagger}$ & 1.5 & 5.8 & 2.4 & 9.9 & 38.9 & 15.8  \\
        DBSCAN++ \cite{najibi2022motion} & 1.4 & 6.1 & 2.2 & 8.9 & 39.9 & 14.5 \\
        DBSCAN++$^{\dagger}$ & 1.6 & 6.2 & 2.5 & 10.2 & 40.3 & 16.2 \\
        DBSCAN++$^{\dagger}_{l}$ & 0.9 & 5.6 & 1.6 & 6.3 & 39.2 & 10.9 \\
        \midrule
        \method$^{\texttt{10}}_{\texttt{sf}}$ & 9.0 & 8.2 & 9.1 & 52.6 & 48.3 & 50.4 \\
        \method$^{\texttt{10}}$ & 10.1 & 8.2 & 9.1 & 59.0 & 48.3 & 53.1 \\
        \method$^{\texttt{50}}$ & 9.0 & 8.9 & 9.0 & 56.7 & 55.7 & 56.1 \\
        \method$^{\texttt{90}}$ &  8.8 & 9.1 & 9.0 & 56.9 & 58.4 & 57.6 \\
    \bottomrule
    \end{tabular}
    \caption{\textbf{Pseudo-label quality comparison (3DIoU):} We compare our \method to different variants of DBSCAN~\cite{najibi2022motion}, augmented with scene flow (DBSCAN++), long-term trajectory information (DBSCAN++$_{l}$) and outlier filtering ($\dagger$). 
    %
    In gray (\textit{top}) we report results using ground truth scene flow and trajectories, and below we report scene flow and motion trajectories obtained using~\cite{wang2023NFT}. 
    %
    %
    \method consistently performs favorably compared to all DBSCAN variants, when using perfect ``oracle'' motion cues, as well as when using the estimated (noisy) scene flow method. 
    } 
    \label{tab:pseudo_quality}
\end{table}

%% file: tables/PLQ_CLASS_WISE_WAYMO.tex
\begin{table}[t]
    \centering
    \footnotesize
    \setlength{\tabcolsep}{5pt}
    \resizebox{0.95\linewidth}{!}{
    \begin{tabular}{l|cc|cc}
    \toprule
         & \multicolumn{2}{c}{DBSCAN++$^{\dagger}$}  &  \multicolumn{2}{c}{\method$^{\texttt{10}}$} \\ 
         &  3DIoU & SegIoU &  3DIoU & SegIoU \\
         & Re (Pr) 0.4 &  Re (Pr) 0.4 & Re (Pr) 0.4 & Re (Pr) 0.4 \\
        \midrule
        Vehicle & 36.7 & 70.1 & 54.5 & 76.9\\ 
        Pedestrian & 46.3 & 67.9 & 41.8 & 55.1 \\ 
        Cyclist & 77.1 & 1.0 & 81.3 & 95.8 \\ 
        \midrule
        Class-agnostic & 40.3 (10.2) & 66.6 (16.8) & 48.3 (59.1) & 66.2 (80.9) \\ 
        \hline\hline
        uFP & \multicolumn{2}{c}{\textcolor{red}{\textbf{72.0}}}  & \multicolumn{2}{c}{\textcolor{red}{\textbf{14.5}}} \\ 
    \bottomrule
    \end{tabular}
   }
    \caption{\textbf{Class-wise evaluation of pseudo-labels:} For class-wise evaluation, we assign GT classes to pseudo-labels that have \textit{any} overlap GT. We additionally report the $\%$ unmatched false positives (uFP), \ie, pseud-labels not matched to any GT box. 
    } 
    \label{tab:class_wise}
\end{table}

%% file: tables/PLQ_CLASS_WISE_ARGO.tex
\begin{table}[t]
    \centering
    \footnotesize
    \setlength{\tabcolsep}{5pt}
    \resizebox{0.95\linewidth}{!}{
    \begin{tabular}{l|cc|cc}
    \toprule
         & \multicolumn{2}{c}{DBSCAN++$^{\dagger}$}  & \multicolumn{2}{c}{\method$^{\texttt{90}}$} \\
         &  3DIoU & SegIoU &  3DIoU & SegIoU \\
         & Re (Pr) 0.4 &  Re (Pr) 0.4 & Re (Pr) 0.4 & Re (Pr) 0.4 \\
        \midrule
        Bicyclist &41.7 & 86.4 & 56.3 & 86.4 \\
        Box Truck & 5.0 & 57.1 & 0 & 44.5 \\
        Bus & 0 & 32.0 & 0.4 & 39.3 \\
        Large Vehicle & 6.9 & 30.1  & 17.1 & 44.4 \\
        Motorcyclist & 89.5 &  89.5  & 100 & 100 \\
        Pedestrian & 29.7 & 59.8  & 42.5 & 57.1\\
        Regular Vehicle & 44.5 & 75.2  & 56.9 & 77.9 \\
        Stroller & 0 &  0 & 0.5  & 0.5 \\
        Truck & 0 & 97.1 & 4.4 & 94.1 \\
        Vehicular Trailer & 3.5 & 14.0  & 0& 14.0 \\
        \midrule
        Class-agnostic & 33.3 (7.5) &  59.3 (13.4) & 45.8 (40.1) & 65.2 (57.0) \\
    \bottomrule
    \end{tabular}
   }
    \caption{\textbf{Cross-dataset generalization:} We evaluate \method, trained on 90\% labeled Waymo Dataset, on Argoverse2 dataset. Note that we never train our approach on Argoverse2. 
    We merge Bicycle and Bicyclist as well as Motorcycle and Motorcyclist since they are not distinguishable by motion.} 
    \label{tab:av2_performance}
\end{table}

%% file: tables/PP_WAYMO.tex
\begin{table*}[thp!]
    \centering
    \footnotesize
        \resizebox{1\linewidth}{!}{
    \begin{tabular}{ll|cc|cccc|cccc}
    \toprule
         & & \% Pseudo  & \% GT & Pr 0.7 &  Re 0.7 &  AP 0.7 & mAP 0.7 & Pr 0.4 &  Re 0.4 &  AP 0.4 &  mAP 0.4\\
         \midrule
        \parbox[t]{2mm}{\multirow{8}{*}{\rotatebox[origin=c]{90}{All (Mov. + stat.)}}} & \textcolor{gray}{Stat. + Mov., class-specific} & \textcolor{gray}{0} & \textcolor{gray}{100} & \textcolor{gray}{35.5} & \textcolor{gray}{55.5} & \textcolor{gray}{--} & \textcolor{gray}{37.1} & \textcolor{gray}{69.7} &  \textcolor{gray}{44.6} & \textcolor{gray}{--} & \textcolor{gray}{80.3} \\
         & \textcolor{gray}{Stat. + Mov., class-agnostic} & \textcolor{gray}{0} & \textcolor{gray}{100} & \textcolor{gray}{31.8} & \textcolor{gray}{41.2} & \textcolor{gray}{36.1} & \textcolor{gray}{--} & \textcolor{gray}{51.2} & \textcolor{gray}{66.5} & \textcolor{gray}{64.7} & \textcolor{gray}{--} \\
         & \textcolor{gray}{Mov-only, class agnostic} & \textcolor{gray}{0} & \textcolor{gray}{100} & \textcolor{gray}{34.4} & \textcolor{gray}{19.9} & \textcolor{gray}{15.1} & \textcolor{gray}{--}  & \textcolor{gray}{63.9} & \textcolor{gray}{37.1} & \textcolor{gray}{35.0}  & \textcolor{gray}{--}  \\
          \cmidrule{2-12}
         & DBSCAN++$^\dagger$ & 90 & 10  & 7.4 & 35.1 & 31.1 & -- & 11.5 & 55.0 & 52.0 & -- \\
         & DBSCAN++$^\dagger$ & 100 & 0 & 0.8 & 3.4 & 0.8 & -- & 5.9 & 25.9 & 14.9 & -- \\
         & \method & 90 & 0  & 3.8 & 3.4 & 1.8 & -- & 26.9 & 24.1 & 19.5 & --  \\
         & \method & 90 & 10  & 25.4 & 35.4 & 31.8 & -- & 40.7 & 56.8 & 54.6 & --  \\
        \midrule
        \midrule
        \parbox[t]{2mm}{\multirow{8}{*}{\rotatebox[origin=c]{90}{Moving only}}} & \textcolor{gray}{Stat. + Mov., class-specific} & \textcolor{gray}{0} & \textcolor{gray}{100} & \textcolor{gray}{30.5} & \textcolor{gray}{34.5} &  \textcolor{gray}{--} & \textcolor{gray}{43.2} & \textcolor{gray}{36.4} & \textcolor{gray}{41.1} & \textcolor{gray}{--} & \textcolor{gray}{85.6} \\
         &  \textcolor{gray}{Stat. + Mov., class-agnostic} & \textcolor{gray}{0} & \textcolor{gray}{100} & \textcolor{gray}{16.1} & \textcolor{gray}{52.8} & \textcolor{gray}{44.8} & \textcolor{gray}{--} & \textcolor{gray}{28.1} & \textcolor{gray}{92.4} & \textcolor{gray}{88.7} & \textcolor{gray}{--} \\
         & \textcolor{gray}{Mov-only, class agnostic} & \textcolor{gray}{0} & \textcolor{gray}{100} & \textcolor{gray}{33.9} & \textcolor{gray}{53.7} & \textcolor{gray}{44.3} & \textcolor{gray}{--} & \textcolor{gray}{57.6} & \textcolor{gray}{91.2} & \textcolor{gray}{89.0} & \textcolor{gray}{--}  \\
         \cmidrule{2-12}
         & DBSCAN++$^\dagger$ & 90 & 10  & 2.0 & 34.5 & 29.8 & -- & 4.1 & 70.1 & 61.0 & --  \\
         & DBSCAN++$^\dagger$ & 100 & 0 & 5.9 & 9.7 & 2.4 & -- & 3.6 & 58.6 & 43.2 & -- \\

         & \method & 90 & 0  & 3.8 & 10.7 & 4.2 & -- & 23.3 & 66.0 & 57.5 & --  \\
         & \method & 90 & 10  & 9.4 & 35.4 & 31.7 & -- & 19.7 & 74.6 & 66.2 & -- \\
         \cmidrule{2-12}
        \cmidrule{2-12}
        & DBSCAN++ \cite{najibi2022motion} & 100 & 0 & -- & -- & -- & -- & -- & -- & -- & 40.4 \\
    \bottomrule
    \end{tabular}
    }
    \caption{\textbf{Semi-supervised 3D object detction on Waymo Open Dataset:} We evaluate models on \textbf{all} (\textit{top}) and \textbf{only moving} (\textit{bottom}) on Waymo Open validation set. 
    \% GT indicates the amount of labeled training data, \% Pseudo indicates the amount of pseudo-labeled data. 
    }
    \label{tab:PointPillars}
    \vspace{+0.2cm}
\end{table*}

%% file: tables/PP_AV2.tex
\begin{table}[thp!]
    \centering
    \footnotesize
        \resizebox{0.95\linewidth}{!}{
    \begin{tabular}{ll|ccHHHH|cccH}
    \toprule
         & & P\% & L\% & Pr 0.7 &  Re 0.7 &  AP 0.7 & mAP 0.7 & Pr 0.4 &  Re 0.4 &  AP 0.4 &  mAP 0.4\\
         \midrule
        \parbox[t]{2mm}{\multirow{2}{*}{\rotatebox[origin=c]{90}{All}}} & Labeled  & 0 & 100 &  25.7 & 19.3 & 13.6 & -- & 52.3 & 39.3 & 35.5 & -- \\
         & \method & 100 & 0  & 1.4 & 3.6 & 1.5 & -- & 12.2 & 30.7 & 22.9 & --   \\
        \midrule
        \midrule
        \parbox[t]{2mm}{\multirow{2}{*}{\rotatebox[origin=c]{90}{Mov.}}} & Labeled & 0 & 100 & 24.9 & 47.3 & 36.9 & -- & 45.1 & 85.5 & 82.4 & -- \\
         & \method & 100 & 0  & 1.4 & 10.8 & 4.3 & -- & 8.0 & 64.7 & 57.6 & -- \\
    \bottomrule
    \end{tabular}
    }
    \caption{\textbf{Cross dataset results:} We train PP detector on ground truth data as well as on pseudo labels generated with \method trained on Waymo Open Dataset.}
    \label{tab:PointPillars_AV2}
\end{table}

%% file: sec/5_conclusion.tex
\vspace{-0.2cm}
\section{Conclusions}
\vspace{-0.2cm}
We introduced \method, a data-driven, class-agnostic approach for pseudo-labeling moving objects in Lidar. We devised \method by revisiting correlation clustering in the context of message passing networks training our model to \textit{learn} to decompose graphs constructed from point clouds. We utilized \method to pseudo-label data, as needed to train object detectors, and demonstrated that our data-driven approach performs favorably compared to prior art, and, more importantly, generalizes across datasets. By making our code, experimental data ad baselines publicly available we hope to inspire future efforts in this field of research. 


%% file: sec/X_suppl.tex
\clearpage
\setcounter{page}{1}
\setcounter{section}{0}

\maketitlesupplementary


\begin{abstract}
In this appendix we first discuss limitations as well as computational costs in \cref{sec:limitations}, detail the datasets we evaluate \method on in \cref{sec:dataset_details}, as well as the computation of our SegIoU metric in \cref{sec:seg_iou}. We then detail our point cloud filtering as well as the impact on \method's performance in \cref{sec:filtering} and give a deeper analysis of our graph construction approaches in \cref{sec:graph_construcion}. Afterwards, we give deeper insights on our cluster post-processing with correlation clustering \cref{sec:postprocessing}, bounding box extraction \cref{sec:bb_extraction}, as well as bounding box inflation \cref{sec:bb_inflation}. We compare the latter with the registration introduced in \cite{najibi2022motion} in \cref{sec:registration} and give details on our implementation of \cite{najibi2022motion} in \cref{sec:dbscan_baselines}. Then, we detail the training of \method in \cref{sec:training_MPN} and PointPillars (PP) in \cref{sec:training_PP}, followed by a deeper discussion on the performance PP trained with different pseudo-labels in \cref{sec:PP_performance_analysis}.

\end{abstract}

\section{Limitations and Computational Cost.} 
\label{sec:limitations}
Since \method relies on pre-processed point clouds and predicted motion as input, the quality of both influences the final performance (see \cref{sec:filtering} of supplementary and Tab. \textcolor{red}{3} of the main paper). For \method itself we observe two main limitations.
(i) Our focus on close, moving PL introduces a bias to PointPillars towards regions where typically moving objects are found which leads to a lower recall (see \cref{tab:PointPillars_LONG} All vs. Moving) since during training correct predictions for which no pseudo-labels exist are penalized. This could be addressed by data augmentation. 
(ii) \method is based on learning to group points, and deriving enclosing modal boxes. Hence, this is different to the amodal ground truth bounding boxes of the datasets under investigation and leads to PointPillars predicting too tight bounding boxes. This issue is especially notable for strict evaluation thresholds (see @$0.7$ Tab. \textcolor{red}{6} in the main paper) and large objects (see Tab. \textcolor{red}{4} and \textcolor{red}{5} in the main paper). It could be addressed by utilizing sequential data to obtain amodal estimates as in \cite{najibi2022motion,zhang2023towards}. However, we observed that especially the approach of \cite{najibi2022motion} suffers from point cloud pre-processing and the quality of the predicted motion (see \cref{sec:registration} of supplementary).

\PARit{Computational costs:} Especially per-timestamp motion optimization ($\sim$7 min, 12GB GPU) increases computational costs. Recently, faster self-supervised methods like \cite{Li2023FastNeuralSceneFlow} allow for speed-ups. To train \method, a single 32GB GPU suffices due to the enormous PC reduction during filtering. For PointPillars we utilize 8x16GB GPUs.

\section{Dataset details}
\label{sec:dataset_details}
As intoduced in Sec. \textcolor{red}{4.1} of the main paper, we ablate and evaluate our method using the large-scale Waymo Open dataset~\cite{sun20CVPR} and only utilize Lidar sensory data. The dataset consists of 6.4 hrs of calibrated image and Lidar sensory data, recorded at 10~Hz in Phoenix, San Francisco, and Mountain View. 

Additionally, we assess the generalization of our method on Argoverse2 dataset~\cite{wilson2021argoverse2} recorded at 10~Hz in six U.S. cities, namely Austin, Detroit, Miami, Pittsburgh, Palo Alto, and Washington, D.C. Argoverse2 provides significantly finer-grained semantic labels (30 object classes). In Tab. \textcolor{red}{5} of the main paper we show the performance on those classes that occur as moving objects in our \texttt{val\_gnn} dataset of Argoverse2. 

Waymo Open dataset and Argoverse2 were recorded with different types of (proprietary) Lidar sensors. While Argoverse2 dataset was recorded with two roof-mounted VLP-32C Lidar sensors, \ie, 64 beams in total, Waymo Open dataset was recorded with five Lidar sensors -- one mid-range Lidar (top) and four short-range Lidar sensors (front, side left, side right, and rear). This leads to a denser representation in close-range. We show in our experiments, that despite the representations being different \method is able to generalize between the datasets.

\section{SegIoU Computation}
\label{sec:seg_iou}
To compute our SegIoU metrics, in each frame we utilize the extracted pseudo-label bounding boxes $b_c \in B_c$ as well as the ground truth bounding boxes $g_c \in G_c$ to find their interior points in the filtered point cloud $\Tilde{P} \in R^{M \times 3}$. This leaves us with binary instance segmentation masks $I_G \in R^{M \times |G_c|}$ and $I_B \in R^{M \times |B_c|}$ for ground truth and pseudo-label bounding boxes, respectively, where the masks indicate point membership to a bounding box. Finally, we compute the intersection over union between $I_B$ and $I_G$ to obtain out SegIoU matrix $S \in R^{|B_c| \times |G_c|}$.

\section{Point Cloud Filtering}
\label{sec:filtering}

In this section, we specify the point cloud filtering we apply before predicting point trajectories introduced in Sec. \textcolor{red}{3.1.1} of the main paper. We then discuss its impact on \method's performance as well as the amodal and modal oracle performance we can obtain with the filtered point cloud.

\PAR{Details on Point Cloud Filtering.}
We follow a similar filtering pipeline as~\cite{najibi2022motion}, where we compute an approximate velocity magnitude for each point via Chamfer Distance between a point cloud at timestamp $t$ and $t+4$ where the point clouds are ego-motion-compensated. We then filter points with $|v^t_i| < 0.2 m/s$. We also remove ground points utilizing a ground plane fitting algorithm~\cite{lee22patchworkPP} and only retain points within a range of $80m$ around the ego vehicle and lower than $4m$. 
On Waymo Open dataset the average full point cloud has 175000 points while the average filtered point cloud has 10500 points. 

\PAR{Impact on \method's Performance.}
The point cloud filtering step is delicate: it impacts the quality of the estimated point trajectories $\tau_i$ -- the more stationary points we filter, the better the point trajectories. On the other hand, by removing too many non-stationary points we may filter out several moving instances and hamper recall. 
The quality of filtering static points can be measured by precision and recall, where a true positive is defined as a static point that we successfully remove, while a false positive is a moving point that we accidentially remove. As in \cite{najibi2022motion} we define static points as points with a ground truth velocity $< 1 m/s$.  Despite closely following the point cloud filtering in \cite{najibi2022motion}, with additional feedback of the authors, we were not able to achieve the recall and precision given in \cite{najibi2022motion} -- they report precision and recall of $97.2\%$ and $62.2\%$ while we achieve $88.3\%$ and $97.5\%$. This leaves us with a more noisy point cloud than used in \cite{najibi2022motion}, where more moving points are filtered out and more static points kept. 

\input{tables/PC_FILTERING}
\PAR{Amodal Oracle Performance of \method}
In Tab.~\ref{tab:filtered_pc} we discuss the number of \textit{static} and \textit{moving} ground truth bounding boxes before (\textit{original}) and after (\textit{filtered}) point cloud filtering with $\geq x_{f}$ interior points in the rectangular region of $100x40m$ around the ego vehicle on the \texttt{val\_pseudo} dataset. We observe that a certain amount of ground truth bounding boxes already have no interior points even in the original point cloud (see difference between $x_f=0$ and $x_f=1$). After filtering, the percentage of static objects is significantly reduced, at the cost of losing some moving objects. However, we retain a high number of boxes of moving objects, \eg, 90\% with at least 10 points, and 67\% with at least 50. This amount of moving objects present in the filtered point cloud determines the \textit{amodal} upper performance bound, \ie, the quality upper bound of our pseudo-labels.

\PAR{Modal Oracle Performance of \method.}
However, our pseudo-labels are \textit{modal}, \ie, the extent of the bounding boxes is defined by the smallest enclosing cuboid around the points present in the filtered point cloud.
To obtain a modal upper bound, we compute an \textit{Oracle} performance where we assign each point its ground truth identity and extract bounding boxes given those ground truth identities. This gives us the performance we would obtain if \method would be able to segment all points correctly. In \cref{tab:pc_filtering_modal} we report this oracle performance utilizing SegIoU- and 3DIoU-based evaluation if we discard segments with $\leq x_f$ interior points. We see that SegIoU performance corresponds to the amodal upper bound -- the slight differences are caused by our bounding box extraction. 3DIoU-based performance shows that even if \method would segment a given point cloud perfectly, comparing our modal bounding boxes to amodal ground truth bounding boxes would lead to a heavy drop in performance compared to amodal oracle performance. We further observe, the larger $x_f$ gets, the lower the recall but the higher the precision and F1 score. This means, that if \method would wrongly classify edges of clusters with a small amount of interior points as negative classes, 3DIoU-based performance would increase. Hence, SegIoU is better suited than 3DIoU to evaluate the clustering quality of \method and we only use 3DIoU-based evaluation to evaluate the final pseudo-label quality and PP performance. 

\begin{table}[t]
    \centering
    \footnotesize
    \setlength{\tabcolsep}{2.5pt}
    \begin{tabular}{l|ccc|ccc} 
    \toprule
        & \multicolumn{3}{c}{\textit{SegIoU}} & \multicolumn{3}{c}{\textit{3DIoU}}\\
        Method & Pr 0.4 & Re 0.4 & F1 0.4 & Pr 0.4 & Re 0.4 & F1 0.4 \\
        \midrule
        $x_f = 2$ & 95.1 & 94.8 & 95.0 & 40.6 & 40.6 & 40.6 \\
        $x_f = 30$ & 97.9 & 81.2 & 88.8  & 48.8 & 40.4 & 44.2   \\
        $x_f = 50$ &  98.1 & 68.8 & 80.8 & 56.5 & 39.6 & 46.6 \\
    \bottomrule
    \end{tabular}
    \caption{
    \textbf{\method$^{\texttt{10}}$ ablation modal upper bound:} We report upper bound via \textbf{oracle}, \ie, the achievable performance with segments with at least $x_f$ interior points from GT labels.
    } 
    \label{tab:pc_filtering_modal}
\end{table}

\begin{table}[t]
    \centering
    \footnotesize
    \setlength{\tabcolsep}{2.5pt}
    \begin{tabular}{l|ccc|ccc} 
    \toprule
        Method & Pr 0.7 & Re 0.7 & F1 0.7 & Pr 0.4 & Re 0.4 & F1 0.4\\
        \midrule
        \textbf{Oracle} Velocity kNN & 35.7 & 67.1 & 46.6 & 39.4 & 74.1 & 51.4 \\
        \textbf{Oracle} Position kNN & 85.0 & 87.9 & 86.4 & 89.4 & 92.5 & 90.9 \\
        \midrule
        Velocity kNN & 35.9 & 41.7 & 38.6 & 45.6 & 52.9 & 49.0   \\
        Position kNN & 63.3 & 53.0 & 57.7 & 77.9 & 65.1 & 70.9  \\
        
    \bottomrule
    \end{tabular}
    \caption{
    \textbf{\method$^{\texttt{10}}$ ablation graph construction (SegIoU):} We discuss the oracle performance of \method utilizing different strategies on \textbf{graph construction} as well as the performance when utilizing those strategies for training.} 
    \label{tab:graph_construction}
\end{table}

\section{Deeper analysis of our graph construction}
\label{sec:graph_construcion}
In the main paper, we report the oracle performance, \ie, the maximum achievable performance given an underlying graph construction strategy when utilizing a kNN graph in Tab. \textcolor{red}{1}. We showed, that the oracle performance of position-based kNN-graph construction is significantly higher than the oracle performance when utilizing velocity. In this section, we discuss those results more in detail as well as the performance when training \method on the different graph construction approaches.

\begin{figure}
\centering
\begin{subfigure}{0.47\linewidth}
  \centering
  \includegraphics[width=\linewidth]{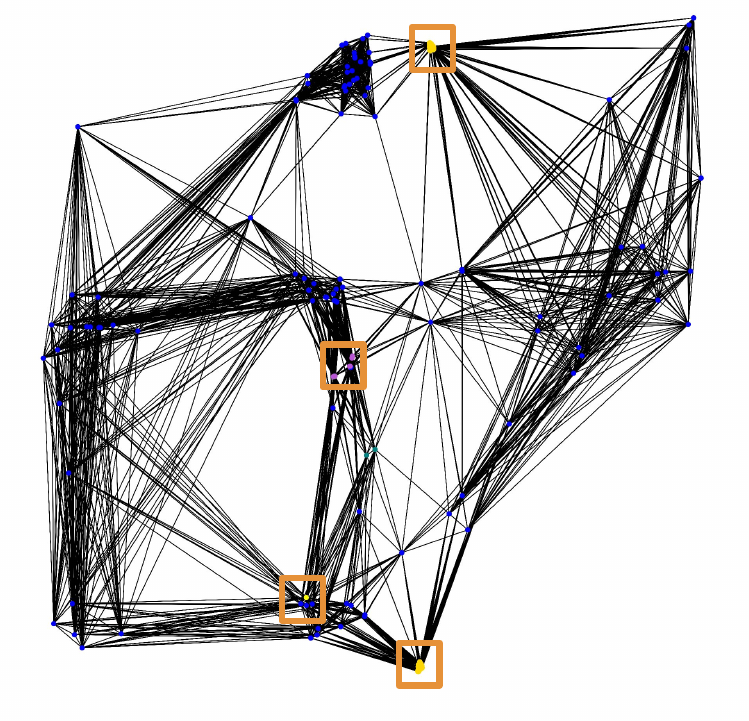}
  \caption{Position-based construction.}
  \label{fig:p_graph}
\end{subfigure}%
\begin{subfigure}{0.47\linewidth}
  \centering
  \includegraphics[width=\linewidth]{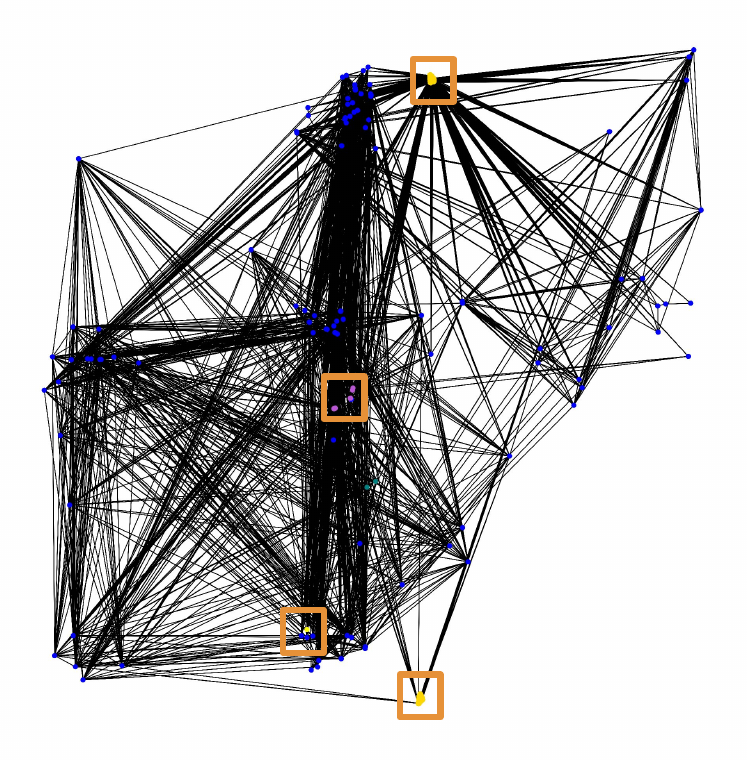}
  \caption{Velocity-based construction.}
  \label{fig:v_graph}
\end{subfigure}
\caption{\textbf{Comparison of different graph construction approaches:} We compare three different graph construction approaches as initial hypothesis for \method: position-based, velocity-based and a combination of both, where we first build a graph based on position and then cut edges if node velocities are highly different. Blue points represent nodes \ie points in the point cloud, edges the initial hypothesis of connection to be refined by our GNN. Position-based graph construction utilizes the inductive-bias of proximity, velocity-based hypothesis yields edges spanning the entire scene since points can potentially have a similar velocity if they are are far in space. 
}
\label{fig:graph_construction}
\end{figure}

\PAR{Oracle Graph Construction Performance.} For velocity-based graph construction, closest points can be situated anywhere in space (see also Fig~\ref{fig:v_graph}) which leads to several negative edges being added to the graph. This is mirrored in the low performance in recall, \ie, many false negative segments, as well as precision, \ie, many false positive segments as reported in Tab. \textcolor{red}{1} in the main paper. Utilizing position as an initial inductive bias assures that close-by points are considered first as belonging to the same object (see Fig~\ref{fig:p_graph}). Hence, computing the oracle performance leads to significantly better performance.

\PAR{\method's Graph Construction Performance.}
Remarkably, utilizing velocity as a graph construction method for \method leads to a higher precision than when computing the oracle performance (see \cref{tab:graph_construction}). This seems counter-intuitive since the oracle performance should lead to the best possible performance achievable. However, the oracle approach leads to scattered objects and more false positive pseudo-labels due to the edge hypothesis' connecting distant points and not connecting close points. During the learning process, \method fails to correctly classify edges of heavily scattered objects leading to them being discarded due to being singleton points or being filtered out by the size filter (any dimension $<0.1m$). This in turn leads to less false positive predictions hence a higher precision but also a lower recall. 
Since (i) this behavior is not desirable during our learning process and (ii) the performance is still significantly worse than when utilizing position-based graph construction, we utilized the latter.

\section{Cluster Postprocessing}
\label{sec:postprocessing}
As introduced in Sec. \textcolor{red}{3.1.2} in the main paper, we obtain edge scores $\Tilde{h}_{i,j}^{(L)}$ for our edges from a binary classifier on the edge features after the last layer. We cut negative edges, \ie, edges for which we predict $\Tilde{h}_{i,j}^{(L)}<0.5$. Then, we remove singleton nodes and apply correlation clustering on the remaining node and edge set utilizing our edge scores as edge weights. For correlation clustering, we utilize the implementation of \cite{abbas2022rama} with default settings.

\section{Bounding Box Extraction}
\label{sec:bb_extraction}
As mentioned in Sec. \textcolor{red}{3.1.1} of the main paper, we discard singleton nodes that are left without edges after our cluster postprocessing and correlation clustering. Additionally to the bounding box extraction defined in Sec. \textcolor{red}{3.2} of the main paper, we discard bounding boxes where either of the dimensions is $< 0.1m$.

\section{Bounding Box Inflation}
\label{sec:bb_inflation}
\method generates compact, modal bounding boxes that are tightly fitted around the clustered points by extracting to smallest enclosing cuboid. 
Due to point cloud filtering and possibly wrongly filtering moving points, pseudo labels are more compact when extracted on the filtered point cloud. However, ground truth bounding boxes are amodal representations of the objects and typically more loosely placed around the object. This leads to the fact that despite \method clustering points correctly, our extracted pseudo-labels are not considered as true positives. Particularly for pedestrians, we observed our pseudo-labels being perfectly fit around the points but being considered as false positive detections. We show several visualizations in \cref{fig:ped_vis} where green represents ground truth bounding boxes and red our pseudo-labels fitted around a point cluster of the filtered point cloud. 

To enhance our pseudo-labels to correspond to the underlying ground truth data to train the detector, we inflate our bounding boxes to have a minimum size of $(1m, 1m, 2m)$ for Waymo Open dataset and $(0.75m, 0.75m, 1.75m)$ for Argoverse2. We show the impact on the performance on Waymo Open Dataset in Tab. \textcolor{red}{2} of the main paper. The performance evaluation based on SegIoU does not change significantly indicating that \method clusters points correctly. When evaluating on 3DIoU, our simple bounding box inflation leads to a drastic performance increase indicating the better correspondence.

\begin{figure}
\centering
  \centering
  \includegraphics[width=0.9\linewidth]{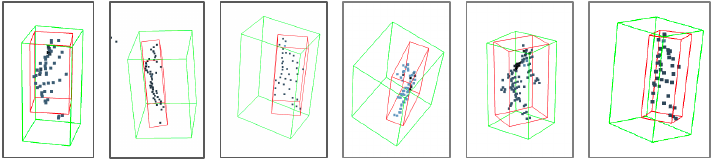}
  \label{fig:ped_vis}
\caption{\textbf{Visualizations of Pedestrian Boudning Boxes:} We show visualizations of pedestrian clusters in our filtered point cloud, our extracted bounding boxes (red) as well as ground truth bounding boxes (green). We can see, that \method clusters points correctly, but the extracted bounding boxes are significantly smaller than their corresponding ground truth bounding boxes.
}
\label{fig:ped_vis}
\end{figure}

\section{Registration}
\label{sec:registration}
For a fair comparison with our baseline DBSCAN++$^{\dagger}$ \cite{najibi2022motion}, we also implemented the registration \cite{gross2019AlignNet} algorithm provided in the paper. The authors state that they track bounding boxes with a BEV IoU threshold of $0.1$, \ie, a small overlap is enough for two bounding boxes to be matched together. Then they register bounding boxes over all tracks to obtain a more complete representation of the underlying object. Contrary to the performance reported in \cite{najibi2022motion}, adding registration to our pipeline worsened our performance (see \cref{tab:ablation_registration}). We assume that this is due to our more noisy point cloud filtering. As soon as one detection in a track is noisy, due to the registration all bounding boxes in a track will be noisy. We also show the performance of our best registration setting, where we: (i) utilize a threshold of $0.5$, (ii) only apply constrained ICP to obtain transformations between detections but propagate the dimensions of the most dense point cloud, (iii) only apply constrained ICP on clusters with more than $50$ points and tracks with a length $\geq 5$. However this also does not improve our performance.
Hence, we rely on bounding box inflation as our only post-processing step due to its simplicity and high efectiveness. 

\input{tables/REGISTRATION}

\section{Details DBSCAN and DBSCAN++ re-implementation}
\label{sec:dbscan_baselines}
For the vanilla DBSCAN baseline based on solely position, we set $\epsilon=1$ and the minimum number of samples per cluster to $10$. As introduced in the main paper, we also compare our approach to the DBSCAN++ approach introduced in \cite{najibi2022motion}. Unfortunately, we were not able to access the implementation as well as the scene flow used and, therefore, report our best re-implementation. To be specific, we apply a DBSCAN clustering on the spatial positions of points in the point cloud and on the scene flow that we extract from our predicted trajectories. Then we utilize the intersection of both to obtain the final clusters. We follow \cite{najibi2022motion} and set $\epsilon_{pos}=1$ and $\epsilon_{flow}=0.1$. To filter out noise, we define the minimum number of samples per cluster to $10$ for both and to $20$ for the intersection. Allowing any number of interior points, \ie, also clusters with a single interior point leads to a significant drop in precision and F1 score -- from $10.2$ to $2.8$ and from $16.2$ to $5.3$ for $T=0.4$ on Waymo Open dataset (see \cref{tab:dbscan_baseline}). Since no minimum number of samples is mentioned in  \cite{najibi2022motion}, we assume that either the point cloud filtering or the scene flow prediction were cleaner to begin with. 
\input{tables/DBSCAN_BASELINE}

\section{\method training.} 
\label{sec:training_MPN}
In this section, we detail the training of \method. As supervisory signal, we define edges as positive, if the two connected nodes belong to the same moving object. Every other edge we consider as negative. This leaves us with the vast majority of edges being negative. Hence, we use the focal loss function~\cite{lin2017focal}, well-suited for our imbalanced binary edge classification objective. For optimization, we use Adam optimizer~\cite{kinga2015adam} and step learning rate schedule with step size 15 and gamma value 0.7 for 30 epochs, batch size of 4 and a base learning rate of 0.003. We apply dense supervision, \ie, we apply the loss function after each layer of the MPN and to all edges. 

\begin{figure}
    \centering
    \includegraphics[width=0.45\textwidth]{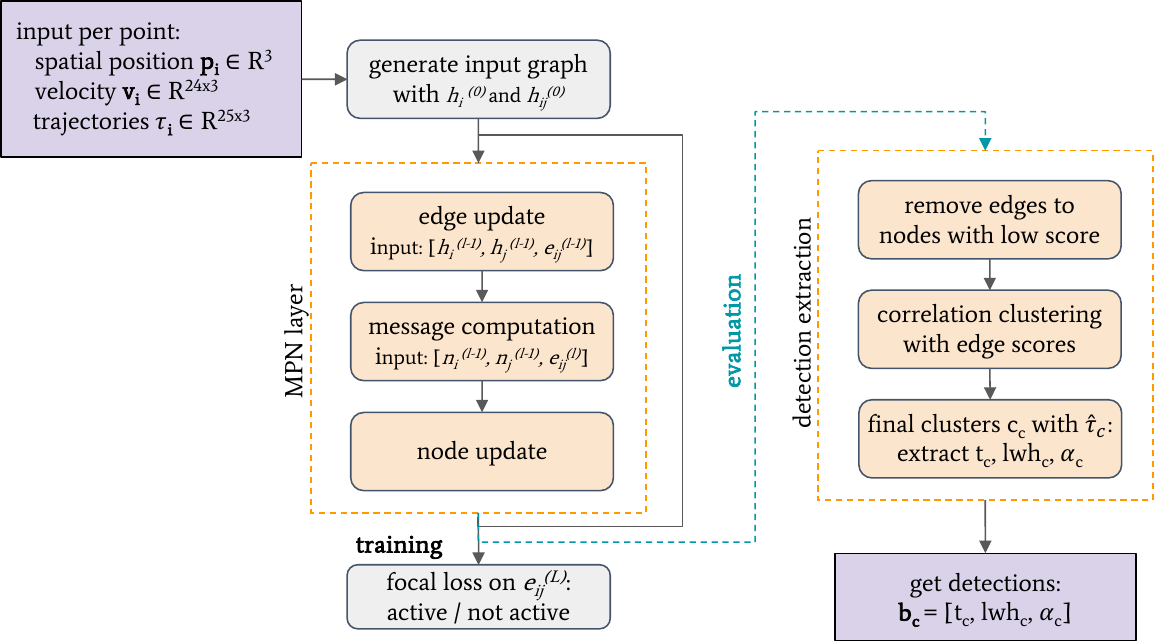}
    \caption{\textbf{Overview MPN during training and evaluation:} We visualize our MPN. It takes as input points from a point cloud with their corresponding spatial positions $p_i$, trajectory $t_i$, and velocities along the trajectory $v_i$. We then extract initial node and edge features $h_i^{(0)}$ and $h_{ij}^{(0)}$, apply $L$ MPN layers and for training apply focal loss on the final edge features $h_{ij}^{(L)}$. During Evaluation, we prune edges based on the final edge scores, apply correlation clustering on the remaining, and extract bounding boxes $b_c$ with translation $t_c$, dimensions $lwh_c$, and heading in $xy$-direction $\alpha_c$.}
    \label{fig:GNN_overview}
\end{figure}

\section{PointPillars Detector Training and Anchor Bounding Boxes}
\label{sec:training_PP}
For our detector training on Waymo Open dataset as well as Argoverse2, we utilize the default implementation and parameters for PointPillars training on Waymo Open dataset of \cite{mmdet3d2020}. To be specific, we train for 24 epochs with a base learning rate of 0.001 and a batch size of 32. We apply a combination of linear and step learning rate decay. For augmentation, we utilize random horizontal and vertical flip, global rotation and scaling as well as point shuffling. We only load every 5\textit{-th} frame. For the Waymo Open dataset we utilize spatial position, intensity, and elongation as input while for Argoverse2 we only use the first two since elongation is not provided. 

To account for different object sizes we utilize three different anchor bounding box sizes on Waymo Open dataset: $(4.75, 2.0, 1.75), (0.9, 0.85, 1.75)$ and $(1.8, 0.85, 1.75)$. Due to the more diverse classes on Argoverse2 dataset, we utilize also a more diverse set of anchor bounding boxes: $(0.75, 0.75, 0.75)$, $(1.5, 0.75, 1.5)$, $(4.5, 2, 1.5)$, $(6.0, 2.5, 2)$, $(9.0, 3.0, 3.0)$, and $(13.0, 3.0, 3.5)$. We obtain those utilizing k-means algorithm on the \texttt{val\_gnn} dataset.

\section{PP Performance with different percentages of pseudo and labeled data}
\label{sec:PP_performance_analysis}
In this section, we detail PointPillars (PP) performance on different percentages of labeled and pseudo-labeled data (see \cref{tab:PointPillars_LONG}). For this we compare different versions of \method, \ie, \method$_{\texttt{10}}$, \method$_{\texttt{50}}$, and \method$_{\texttt{90}}$ with DBSCAN++$^{\dagger}$ and our baselines trained on ground truth data. 

\PAR{Comparison of Different Versions of \method.}
We first compare the PP performance obtained when utilizing the different versions of \method but the same data set split of pseudo-labeled data. To be specific, we utilize the \texttt{train\_detector} set containing $10\%$ of our training data (please refer to Fig. \textcolor{red}{3} of the main paper for split definition). 

\PARit{Evaluation on static and moving objects.}
In \cref{tab:PointPillars_LONG} we show that the overall performance of \method$_{\texttt{10}}$ falls behind the two other versions. This shows that more training data indeed improves the performance of \method. However, a training split larger than $x=50\%$ does not seem to improve the overall performance further when evaluating the performance with $T=0.4$. For the more strict threshold evaluation of $T=0.7$, we can observe slightly favorable behavior of \method$_{\texttt{90}}$. Interestingly, the precision of \method$_{\texttt{50}}$ falls behind the other models.

\PARit{Evaluation on moving objects.}
The same observations hold when evaluating on moving objects only with all observations being more distinct. This shows that \method only extracts moving objects.

\PAR{Training without Labeled Data.}  
We compare the performance when we train PP with pseudo-labels generated using DBSCAN++$^{\dagger}$ and the different versions of \method on different amounts of pseudo-labeled data. 

\PARit{Evaluation on static and moving objects.}
Interestingly, for \method as well as for DBSCAN++$^{\dagger}$ without labeled data, the best performance is achieved utilizing $50\%$ of the pseudo-labeled data. Precision as well as recall are always similar forDBSCAN++$^{\dagger}$ with a constantly low precision. This indicates that the noisy training signal leads to PP predicting a large number of false positive bounding boxes.
Pseudo-labels generated by \method lead to increased precision as we increase the amount of pseudo-labeled data mirroring the high pseudo-label quality. However, recall drops which indicates that pseudo-labels generated by \method indeed focus on moving objects. The higher the ratio of moving objects PP obtains as supervisory signal the less static objects it will predict.

\PARit{Evaluation on moving objects.}
On moving objects only, the performance of DBSCAN++$^\dagger$ when utilizing only $10\%$ pseudo-labeled data is significantly worse than the performances utilizing $50\%, 90\%$ or $100\%$ which are relatively stable. This indicates that DBSCAN++$^\dagger$ pseudo-labels in general contain many noisy pseudo labels not belonging to moving objects and that the amount of moving objects is not enough in the $10\%$ split. Afterwards the DBSCAN++$^\dagger$ pseudo-labels with low singal-to-noise ratio do not add to the training signal. For \method, we again observe a constant increase of precision the more pseudo-labeled data we utilize, indicating that the overall higher quality of the bounding boxes can be transferred to training PP. 

\PAR{Training with Labeled Data.}  
Finally, we compare the performance when we train PP on a combination of labeled data and pseudo-labeled data generated using DBSCAN++$^{\dagger}$ and the different versions of \method on different data splits. 

\PARit{Evaluation on static and moving objects.}
\method profits significantly more from adding labeled data to PP training with respect to precision than DBSCAN++$^\dagger$. This indicates that the pseudo-labels generated by DBSCAN++$^\dagger$ hamper the prediction of precise pseudo labels compared to when using \method's pseudo-labels. This hypothesis is supported by the fact that the precision using DBSCAN++$^\dagger$ does not significantly change depending on the amount of labeled data, while \method shows the highest precision when using the largest amount of labeled data. Since DBSCAN++$^\dagger$ generates noisy pseudo-labels with many predictions not belonging to moving objects, the recall is highly similar to the recall when utilizing \method. This leads to an overall not significantly worse AP value of DBSCAN++$^\dagger$. 

\PARit{Evaluation on moving objects.}
Evaluating on moving objects only leads to similar observations, except that the recall when utilizing \method's pseudo-labels is higher compared to when utilizing DBSCAN++$^\dagger$ pseudo-labels. This again indicates that indeed \method's pseudo-labels focus on moving objects while DBSCAN++$^\dagger$'s pseudo-labels also contain many pseudo-labels not belonging to moving objects -- the vast majority belonging to noise. 

\input{tables/PP_WAYMO_LONG}

%% file: tables/PC_FILTERING.tex
\begin{table}[ht]
    \centering

    \footnotesize
    \setlength{\tabcolsep}{5pt}
    \begin{tabular}{l|cc|cc}
    \toprule
         & \multicolumn{2}{c}{\textit{static objects}}  & \multicolumn{2}{c}{\textit{moving objects}}\\
         
        \# Interior & original & filtered & original & filtered \\
        \midrule
        $x_f$ = 0 & 100 & 100 & 100 & 100 \\
        $x_f$ = 1 & 97.8 & 71.1 & 96.3 & 95.4 \\
        $x_f$ = 10 & 94.9 & 21.8 & 92.4 & 90.4 \\
        $x_f$ = 30 & 89.5 & 9.7 & 84.3 & 79.5 \\
        $x_f$ = 50 & 82.6 & 6.3 & 76.2 & 67.7 \\
        
    \bottomrule
    \end{tabular}
    \caption{\textbf{Point cloud filtering.} In the pre-filtering step, we remove points that appear static. We report the percentage of labeled bounding boxes that have more $\geq x_f$ interior before and after filtering. A few observations: (i) even when not filtering the point cloud, a certain percentage of boxes have no interior points; those are probably occluded and extrapolated. (ii) filtering causes a big drop for static objects, as expected. (iii) after filtering we retain a high number of boxes of moving objects: 90\% with at least 10 points, and 67\% with at least 50. Those are those that we can obtain in the pseudo-labeling process.} 
    \label{tab:filtered_pc}
\end{table}

%% file: tables/REGISTRATION.tex
\begin{table}[t]
    \centering
    \resizebox{1.0\linewidth}{!}{
    \footnotesize
    \setlength{\tabcolsep}{5pt}
    \begin{tabular}{l|ccc||ccc}
    \toprule
        & \multicolumn{3}{c}{\textit{3DIoU}} & \multicolumn{3}{c}{\textit{SegIoU}} \\
          & Pr 0.4 & Re 0.4 & F1 0.4 & Pr 0.4 & Re 0.4 & F1 0.4 \\
        \midrule
        Initial & 33.2 & 27.8 & 30.3 & 77.9 & 65.1 & 70.9 \\
        Registration as in \cite{najibi2022motion} & 2.1 & 0.9 & 1.2 & 20.5 & 8.6 & 12.2 \\
        Our best Registration & 21.0 & 15.0 & 17.5 & 67.8 & 48.4 & 56.5 \\
        Inflated & 59.1 & 48.3 & 53.1 & 80.9 & 66.2 & 72.8\\
    \bottomrule
    \end{tabular}
   }
    \caption{\textbf{Registration:} We implement constrained ICP \cite{gross2019AlignNet} for bounding box amodalization as in \cite{najibi2022motion}. Due to our more noisy point cloud and trajectory predictions we are not able to improve our performance. However, simple bounding box inflation leads to a significant performance improvement.} 
    \label{tab:ablation_registration}
\end{table}

%% file: tables/DBSCAN_BASELINE.tex
\begin{table}[t]
    \centering
    \resizebox{1.0\linewidth}{!}{
    \footnotesize
    \setlength{\tabcolsep}{5pt}
    \begin{tabular}{l|ccc||ccc}
    \toprule
          & Pr 0.7 & Re 0.7 & F1 0.7 & Pr 0.4 & Re 0.4 & F1 0.4 \\
        \midrule
        Re-implementation as in \cite{najibi2022motion} & 0.4 & 5.2 & 0.7 &  2.8 & 37.7 & 5.3 \\
        Adapted re-implementation & 1.6 & 6.2 & 2.5 & 10.2 & 40.3 & 16.2 \\
    \bottomrule
    \end{tabular}
   }
    \caption{\textbf{Comparison of DBSCAN++$^{\dagger}$ with and without minimum number of points per cluster:} We compare the performance of the re-implementation of our baseline DBSCAN++$^{\dagger}$ as described in \cite{najibi2022motion} where clusters are allowed to contain only one point. In our adapted re-implementation we define the minimum number of points per cluster for position- as well as scene flow-based clusterings as $10$ and the minimum number of samples of their intersection as $20$.} 
    \label{tab:dbscan_baseline}
\end{table}

%% file: tables/PP_WAYMO_LONG.tex
\begin{table*}[thp!]
    \centering
    \footnotesize
        \resizebox{1\linewidth}{!}{
    \begin{tabular}{ll|cc|cccc|cccc}
    \toprule
         & & \% Pseudo  & \% GT & Pr 0.7 &  Re 0.7 &  AP 0.7 & mAP 0.7 & Pr 0.4 &  Re 0.4 &  AP 0.4 &  mAP 0.4\\
         \midrule
        \parbox[t]{2mm}{\multirow{19}{*}{\rotatebox[origin=c]{90}{All (Mov. + stat.)}}} 
        & \multicolumn{11}{l}{\textit{Ground Truth Baselines}} \\
        & \textcolor{gray}{Stat. + Mov., class-specific} & \textcolor{gray}{0} & \textcolor{gray}{100} & \textcolor{gray}{35.5} & \textcolor{gray}{55.5} & \textcolor{gray}{--} & \textcolor{gray}{37.1} & \textcolor{gray}{69.7} &  \textcolor{gray}{44.6} & \textcolor{gray}{--} & \textcolor{gray}{80.3} \\
         & \textcolor{gray}{Stat. + Mov., class-agnostic} & \textcolor{gray}{0} & \textcolor{gray}{100} & \textcolor{gray}{31.8} & \textcolor{gray}{41.2} & \textcolor{gray}{36.1} & \textcolor{gray}{--} & \textcolor{gray}{51.2} & \textcolor{gray}{66.5} & \textcolor{gray}{64.7} & \textcolor{gray}{--} \\
         & \textcolor{gray}{Mov-only, class agnostic} & \textcolor{gray}{0} & \textcolor{gray}{100} & \textcolor{gray}{34.4} & \textcolor{gray}{19.9} & \textcolor{gray}{15.1} & \textcolor{gray}{--}  & \textcolor{gray}{63.9} & \textcolor{gray}{37.1} & \textcolor{gray}{35.0}  & \textcolor{gray}{--}  \\
          \cline{2-12}
          \cmidrule{2-12}
          & \multicolumn{11}{l}{\textit{Different versions of} \method} \\
         & \method$_{\texttt{90}}$ & 10 & 0  & 3.1 & 3.2 & 1.5 & -- & 24.0 & 24.8 & 19.6 & -- \\
         & \method$_{\texttt{50}}$ & 10 & 0  & 2.8 & 3.7 & 1.9 & -- & 21.1 & 27.7 & 21.8 & --\\
         & \method$_{\texttt{10}}$ & 10 & 0  & 3.1 & 3.8 & 1.9 & -- & 22.8 & 27.3 & 21.7 & -- \\
         \cline{2-12}
         \cmidrule{2-12}
         
         & \multicolumn{11}{l}{\textit{Training with pseudo-labeled data only}} \\
          & DBSCAN++$^\dagger$ & 10 & 0  & 0.6 & 3.0 & 0.8 & -- & 5.1 & 26.0 & 14.2 & --  \\
          & DBSCAN++$^\dagger$  & 50 & 0  & 0.7 & 3.0 & 0.7 & -- & 5.8 & 25.9 & 15.3 & --  \\
          & DBSCAN++$^\dagger$ & 90 & 0  & 0.7 & 3.2 & 1.2 & -- & 5.8 & 25.2 & 14.7 & --\\
          & DBSCAN++$^\dagger$ & 100 & 0 & 0.8 & 3.4 & 0.8 & -- & 5.9 & 25.9 & 14.9 & -- \\
          \cline{2-12}
         & \method$_{\texttt{90}}$ & 10 & 0  & 3.0 & 3.8 & 1.8 & -- & 21.6 & 27.6 & 21.5 & --  \\
         & \method$_{\texttt{50}}$ & 50 & 0  & 3.7 & 4.1 & 2.0 & -- & 24.6 & 27.6 & 22.6 & -- \\
         & \method$_{\texttt{10}}$ & 90 & 0  & 3.8 & 3.4 & 1.8 & -- & 26.9 & 24.1 & 19.5 & --  \\
         \cline{2-12}
         \cmidrule{2-12}
         
         & \multicolumn{11}{l}{\textit{Training with pseudo-labeled and labeled data}} \\
         & DBSCAN++$^\dagger$ & 10 & 90  & 7.2 & 35.2 &  31.3 & -- & 11.4 & 55.2 & 52.2 & -- \\
         & DBSCAN++$^\dagger$  & 50 & 50  & 6.9 & 35.1 & 31.2 & -- & 10.9 & 54.9 & 51.7 & --  \\
         & DBSCAN++$^\dagger$ & 90 & 10  & 7.4 & 35.1 & 31.1 & -- & 11.5 & 55.0 & 52.0 & -- \\
         \cline{2-12}
         & \method$_{\texttt{90}}$ & 10 & 90  & 49.0 & 35.8 & 32.4 & -- & 70.5 & 51.5 & 50.4 & --  \\ 
         & \method$_{\texttt{50}}$ & 50 & 50  & 29.4 & 36.0 & 32.3 & -- & 46.9 & 57.4 & 55.0 & --  \\
         & \method$_{\texttt{10}}$ & 90 & 10  & 25.4 & 35.4 & 31.8 & -- & 40.7 & 56.8 & 54.6 & --  \\
        \hline
        \midrule

        \parbox[t]{2mm}{\multirow{20}{*}{\rotatebox[origin=c]{90}{Moving only}}} & \multicolumn{11}{l}{\textit{Ground Truth Baselines}} \\
        & \textcolor{gray}{Stat. + Mov., class-specific} & \textcolor{gray}{0} & \textcolor{gray}{100} & \textcolor{gray}{30.5} & \textcolor{gray}{34.5} &  \textcolor{gray}{--} & \textcolor{gray}{43.2} & \textcolor{gray}{36.4} & \textcolor{gray}{41.1} & \textcolor{gray}{--} & \textcolor{gray}{85.6} \\
         &  \textcolor{gray}{Stat. + Mov., class-agnostic} & \textcolor{gray}{0} & \textcolor{gray}{100} & \textcolor{gray}{16.1} & \textcolor{gray}{52.8} & \textcolor{gray}{44.8} & \textcolor{gray}{--} & \textcolor{gray}{28.1} & \textcolor{gray}{92.4} & \textcolor{gray}{88.7} & \textcolor{gray}{--} \\
         & \textcolor{gray}{Mov-only, class agnostic} & \textcolor{gray}{0} & \textcolor{gray}{100} & \textcolor{gray}{33.9} & \textcolor{gray}{53.7} & \textcolor{gray}{44.3} & \textcolor{gray}{--} & \textcolor{gray}{57.6} & \textcolor{gray}{91.2} & \textcolor{gray}{89.0} & \textcolor{gray}{--}  \\
         \cline{2-12}
          \cmidrule{2-12}
          & \multicolumn{11}{l}{\textit{Different versions of} \method} \\
          & \method$_{\texttt{90}}$ & 10 & 0  & 2.9 & 9.6 & 3.4 & -- & 19.4 & 64.2 & 54.3 & -- \\
         & \method$_{\texttt{50}}$ & 10 & 0  & 2.6 & 11.0 & 4.4 & -- & 16.1 & 68.3 & 59.2 & -- \\
         & \method$_{\texttt{10}}$ & 10 & 0  & 2.8 & 10.6 & 4.5 & -- & 17.6 & 67.4 & 58.5 & -- \\
         \cline{2-12}
         \cmidrule{2-12}

         & \multicolumn{11}{l}{\textit{Training with pseudo-labeled data only}} \\
          & DBSCAN++$^\dagger$ & 10 & 0 & 0.5 & 8.7 & 2.4 & -- & 3.0 & 56.4 & 39.7 & --  \\
          & DBSCAN++$^\dagger$  & 50 & 0  & 0.6 & 9.2 & 2.1 & -- & 3.5 & 58.7 & 44.2 & --  \\
          & DBSCAN++$^\dagger$ & 90 & 0  & 0.6 & 9.5 & 1.1 & -- & 3.6 & 58.4 & 43.5 & --  \\
          & DBSCAN++$^\dagger$ & 100 & 0 & 5.9 & 9.7 & 2.4 & -- & 3.6 & 58.6 & 43.2 & -- \\
          \cline{2-12}
          & \method$_{\texttt{90}}$ & 10 & 0  & 2.6 & 10.7 & 4.3 & -- & 17.6 & 67.7 & 58.5 & -- \\
         & \method$_{\texttt{50}}$ & 50 & 0  & 3.4 & 12.1  & 4.8 & -- & 19.7 & 69.9 & 62.7 & -- \\
         & \method$_{\texttt{10}}$ & 90 & 0  & 3.8 & 10.7 & 4.2 & -- & 23.3 & 66.0 & 57.5 & --  \\
         \cline{2-12}
         \cmidrule{2-12}

         & \multicolumn{11}{l}{\textit{Training with pseudo-labeled and labeled data}} \\
         & DBSCAN++$^\dagger$ & 10 & 90  &  2.0 & 34.5 & 30.1 & -- & 4.0 & 70.5 & 61.6 & -- \\
         & DBSCAN++$^\dagger$  & 50 & 50  & 1.9 & 34.3 & 29.6 & -- & 3.8 & 70.2 & 60.2 & -- \\
         & DBSCAN++$^\dagger$ & 90 & 10  & 2.0 & 34.5 & 29.8 & -- & 4.1 & 70.1 & 61.0 & --  \\         
        \cline{2-12}
         & \method$_{\texttt{90}}$ & 10 & 90  & 26.2 & 35.4 & 32.3 & -- & 45.5 & 61.4 & 53.3 & -- \\ 
         & \method$_{\texttt{50}}$ & 50 & 50  & 11.6 & 36.1 & 32.5 & -- & 24.3 & 75.4 & 66.7 & -- \\
         & \method$_{\texttt{10}}$ & 90 & 10  & 9.4 & 35.4 & 31.7 & -- & 19.7 & 74.6 & 66.2 & -- \\
         \cline{2-12}
        \cmidrule{2-12}
        & DBSCAN++ \cite{najibi2022motion} & 100 & 0 & -- & -- & -- & -- & -- & -- & -- & 40.4 \\
    \bottomrule
    \end{tabular}
    }
    \caption{\textbf{Semi-supervised 3D object detection on Waymo Open Dataset:} We evaluate models on \textbf{all} (\textit{top}) and \textbf{only moving} (\textit{bottom}) on Waymo Open validation set. 
    For each evaluation procedure we (i) show the ground truth baselines. Then we (ii) compare the different versions of \method utilizing the same \texttt{train\_detector} training split. To compare \method to our baseline DBSCAN++$^\dagger$ we (iii) show a comparison of both utilizing different percentages of pseudo-labeled data. Finally, we (iv) also show a comparison utilizing different combinations of pseudo-labeled and labeled data.
    \% GT indicates the amount of labeled training data, \% Pseudo indicates the amount of pseudo-labeled data. 
    }
    \vspace{2cm}
    \label{tab:PointPillars_LONG}
\end{table*}